\documentclass[runningheads]{llncs}

\usepackage{eccv}

\usepackage{eccvabbrv}

\usepackage{graphicx}
\usepackage{booktabs}
\usepackage{multirow}
\usepackage{bm}
\usepackage{colortbl}
\usepackage[ruled,linesnumbered]{algorithm2e}
\usepackage{setspace}

\usepackage[accsupp]{axessibility}  % Improves PDF readability for those with disabilities.

\usepackage[table]{xcolor}
\newcommand{\extra}[1]{\textcolor{gray}{\,(#1)}}

\newcommand{\best}[1]{\cellcolor{red!15}\text{#1}}
\newcommand{\second}[1]{\cellcolor{orange!18}\text{#1}}
\newcommand{\third}[1]{\cellcolor{yellow!20}\text{#1}}

\usepackage{hyperref}

\usepackage{orcidlink}

\begin{document}

\title{Color Pass-Through via Camera-Display Coupling} 

\titlerunning{Color Pass-Through}

\author{Ruikang Li\inst{1}\orcidlink{0009-0003-2608-2673} \and
Molin Li\inst{2}\orcidlink{0009-0007-8067-285X} \and
Jiarui Wu\inst{1} \and
Zhe Wei\inst{3} \and
Pengpeng Liu\inst{3} \and
Tianfan Xue\inst{1,\dagger}\orcidlink{0000-0001-5031-6618}}

\authorrunning{R.~Li et al.}

\institute{CUHK MMLab \and Zhejiang University \and
Central Media Technology Institute, Huawei}

\maketitle

\begin{abstract}
    When a real-world scene is captured by a smartphone camera and viewed on its screen, the displayed image often differs noticeably from the original scene in color, brightness, and contrast. 
    This gap persists despite substantial advances in both modern cameras and displays. A key reason is that most pipelines factor the high-dimensional capture-to-display process into two separately calibrated camera and display stages, and then connect them through low-dimensional color transforms, leading to information bottlenecks and inevitable error accumulation. 
    To address this systemic challenge, we propose \textbf{Color Pass-Through}, an end-to-end learned framework that operates directly on captured images. Our key insight is to treat the camera and display as a coupled system rather than calibrating them in isolation.
    Coupling the camera and display yields two practical advantages: (1) it brings the entire real-world scenes to the display via end-to-end optimization, and (2) it allows efficient one-step calibration for each distinct observer via complete capture-to-display path. We validate \textbf{Color Pass-Through} using both digital and human observers. 
    Compared with representative baselines, our method achieves an average gain of $\mathbf{+2.0}$ points on a 5-point user-study and more than $\mathbf{2\times}$ improvement on quantitative metrics, demonstrating improved reproduction of the perceived color of the original scene. 
    See project page: \url{https://lyricccco.github.io/color-pass-through/}
\end{abstract}

\begin{figure*}
  \centering
    \includegraphics[width=\linewidth]{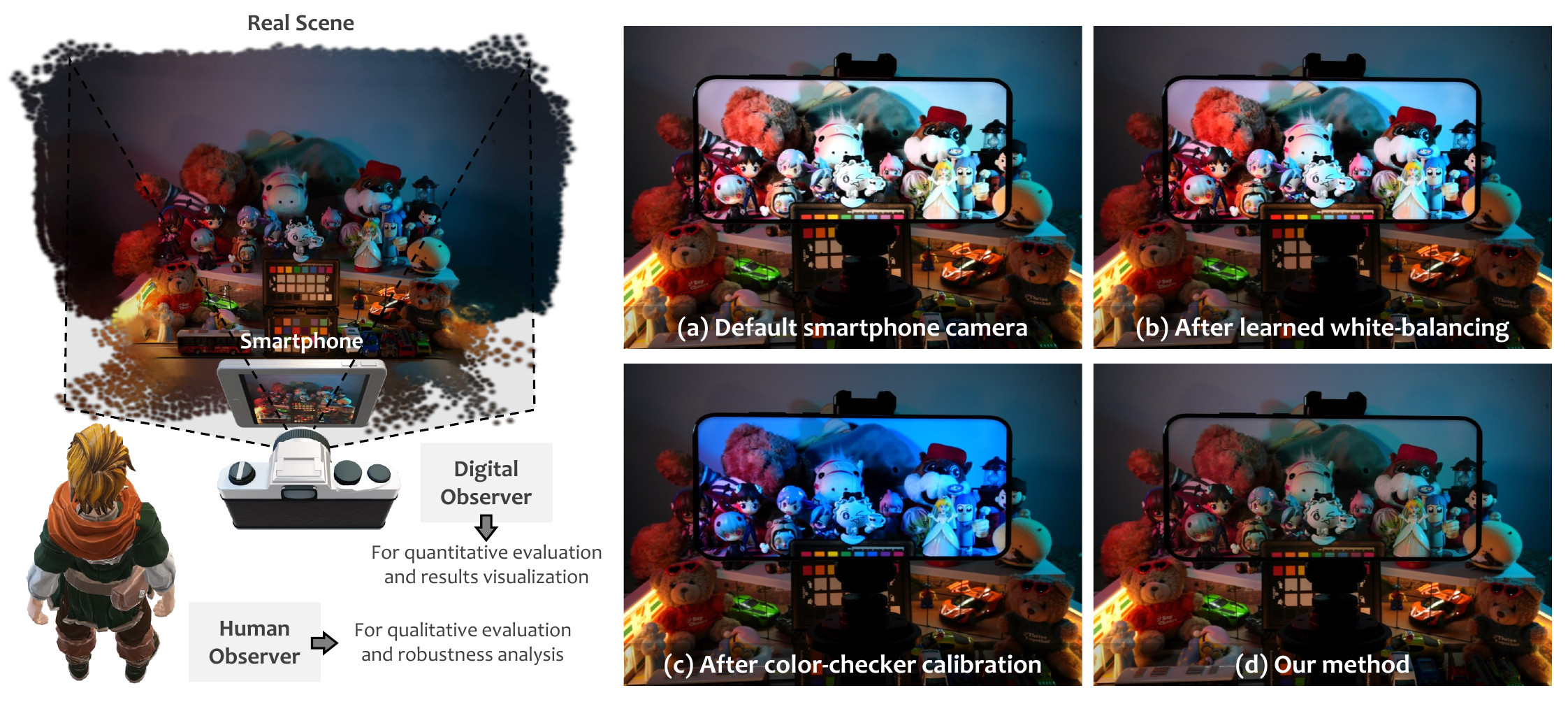}
    \caption{\textbf{Color Pass-Through.} Capturing a real scene under multiple unknown light sources and re-displaying it on a smartphone screen often introduces perceptual color inconsistencies (a). Learned multi-illuminant auto white balance reduces illumination-induced color casts (b), while per-scene color-checker calibration maps the image into a standard color space (c); but neither reproduces the same color. In contrast, our method better preserves the perceived scene colors for both digital and human observers (d).
    }
    \label{fig:teaser}
\end{figure*}

\section{Introduction}
\label{sec_1-Introduction}

Photography seeks to faithfully reproduce a real scene's colors when showing them on a display. In practice, however, what we see in the real world often differs from what a smartphone screen presents. As shown in~\cref{fig:teaser}~(a), capturing a scene with a smartphone and viewing it on the screen can introduce shifts in both chromaticity and lightness: the dolls’ hues drift, and the displayed image may appear overly bright, desaturated, or washed out compared with the original scene. Historically, this perceptual gap was exacerbated by limited sensor dynamic range and narrow-gamut, low-bit panels. Although modern sensors and high-quality displays mitigate this gap, a noticeable discrepancy remains under standard sensor calibration and downstream image post-processing. However, even strong post-processing baselines do not close this gap. Learned multi-illuminant auto white-balance can reduce illumination-induced color casts~\cite{afifi2022awb} (\cref{fig:teaser}~(b)), and per-scene color-checker calibration can map the captured image into a standard color space~\cite{sunoj2018colorcalibration} (\cref{fig:teaser}~(c)).
Fundamentally, these methods rely on constrained assumptions about illumination or standard color space, and therefore can not provide a consistent guarantee of faithful color reproduction for a specific camera–display pair across diverse, unconstrained real-world scenes. This gap is even more problematic in immersive systems: see-through views in VR headsets (e.g., Vision Pro) can deviate significantly from the naked eye in color and contrast, undermining comfort and presence~\cite{bailenson2024seeing,de2024visual}, and virtual try-on mirrors in consumer AR displays have similar fidelity issues~\cite{wang2024perspective}.

\begin{figure}[h]
\centering
  \includegraphics[width=\linewidth]{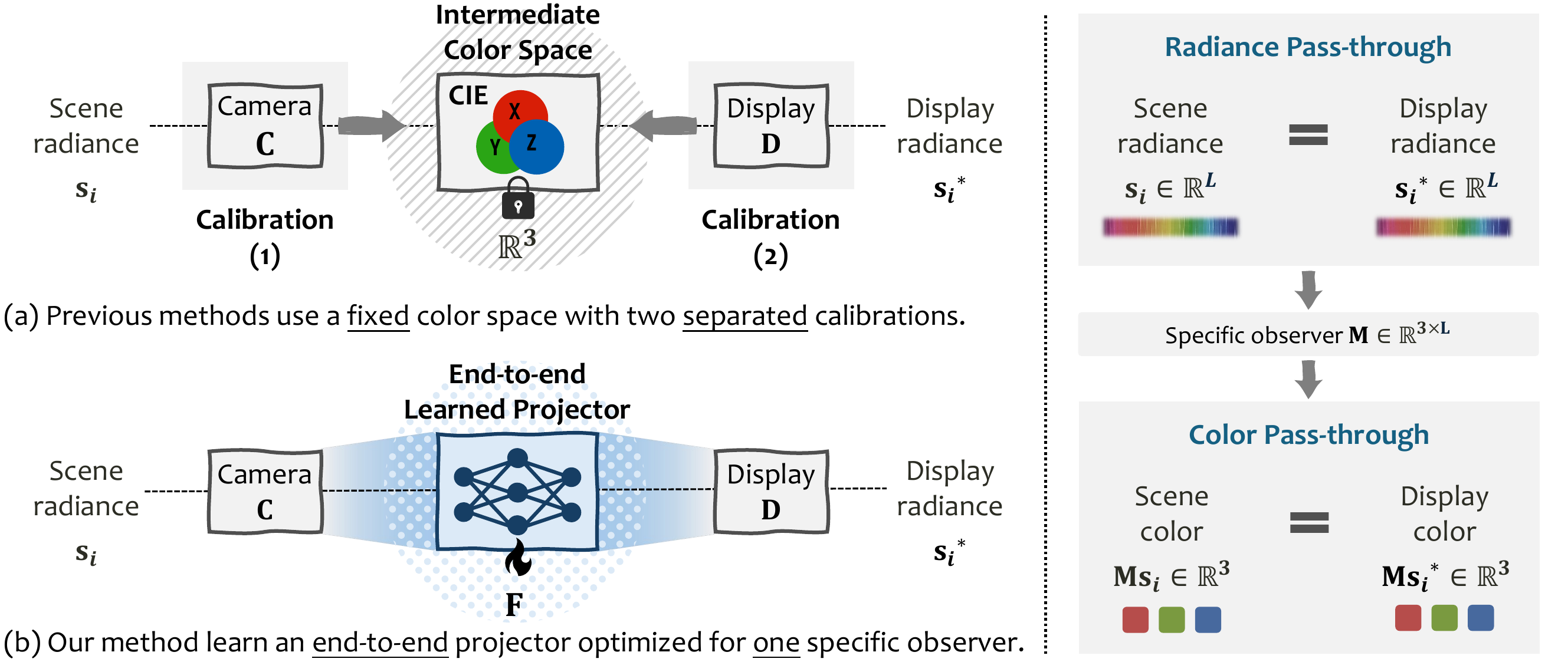}
  \caption{\textbf{Comparison of Methods and Objectives for Color Reproduction.} 
  }
  \label{fig_2}
\end{figure}

To understand why existing solutions struggle to close this color gap, we revisit the capture-to-display imaging pipeline. Standardized International Color Consortium (ICC) workflows~\cite{ICC:2022} rely on a three-channel color representation as a tractable intermediate (\cref{fig_2}~(a)). In doing so, they decompose the process into two separately calibrated stages: \emph{camera calibration}, mapping scene radiance to ``RGB'', and \emph{display calibration}, mapping ``RGB'' to emitted radiance. This separation accumulates error. More fundamentally, camera measurements are inherently three-dimensional, whereas real-world radiance is high-dimensional, so calibration alone cannot overcome this intrinsic information bottleneck.

To close this perceptual color gap, we propose to treat camera and display as a single, coupled system, bridged by an end-to-end learnable neural projector (\cref{fig_2}~(b)). Instead of calibrating each device to a reference color and composing the separate mappings, we directly map displayed colors to the original scene colors for a specific observer. This end-to-end formulation reduces error accumulation and lessens the impact of information bottleneck, yielding markedly smaller mismatches in both color and brightness (\cref{fig:teaser}~(d)). While this may appear to increase the calibration burden, since each camera--display pair must be characterized, it is well suited to pass-through use cases, where users typically view the captured scene on the same phone or immersive device. We therefore calibrate per camera--display pair rather than for every possible combination.

We represent the coupled camera–display pair as an unknown end-to-end non-linear projector (a device-specific mapping), learned with a lightweight pixel-wise neural network. To acquire training data, we introduce a re-capture protocol: each digital RGB sample is rendered on the target display and re-imaged by the paired camera, and the resulting measurements supervise the camera–display projector. We further study dataset design and network architectures, identifying an efficient configuration that learns the projector with a compact model, enabling practical camera–display projection with fast inference.

Despite this learned projector, visible color casts still persist for a target human observer under complex illumination. This residual mismatch arises from camera metamerism. We therefore derive objectives that transition from idealized \emph{radiance pass-through} to \emph{color pass-through} (\cref{fig_2}, right-side), revealing that the dominant discrepancy lies in the camera’s metameric-black subspace: spectral components invisible to the camera can still affect the observer’s perceived color after display, i.e., colors appear identical to the camera may looks totally different to human observers. This limitation is fundamental, because camera spectral sensitivities generally differ from human visual responses.

Empirically, we find that the camera-null space exhibit low intrinsic dimensionality in practice and can be well approximated by a single dominant component, which also keeps the observer-specific correction low-dimensional. We therefore compensate the residual color cast using a learned predictor together with a single observer-specific calibration coefficient $\in\mathbb{R}^{3\times1}$ applied in one step.

\noindent\textbf{Contributions.} This paper propose an end-to-end optimizing system that couples a camera–display pair to achieve color pass-through for a specific observer.

\begin{itemize}
    \item \emph{Learned Camera-Display Projection.} Our core contribution is an efficient pixel-wise neural projector that models the end-to-end mapping of a coupled camera–display, along with a practical re-capture protocol for data collection.
    \item \emph{Camera-Null Color Correction.} We identify the dominant residual error as lying in the camera’s metameric-black (camera-null) subspace, and estimated its main component through a learned predictor with a one-step, observer-specific calibration coefficient to compensate the remaining color cast.
    \item \emph{Experimental Validation.} We validate our method with both objective measurements using a DSLR (a fixed digital observer used to produce all quantitative results and figures) and subjective evaluation via user studies, demonstrating robust color pass-through across diverse scenes and illuminants.
\end{itemize}

\section{Related Work}
\label{sec_2-prior_work}

Traditional color reproduction pipelines (e.g., ICC workflows) rely on an intermediate device-independent reference space such as CIE XYZ to mediate device-to-device transforms.
In this paradigm, capture devices (e.g., cameras and scanners) map sensor measurements into the reference space, while output devices (e.g., displays and printers) apply an inverse mapping to device-dependent signals to reproduce the intended colors. 
This design yields two decoupled calibration stages: \emph{camera calibration} and \emph{display calibration}. We focus on the lines of work most relevant to our setting, and refer readers to~\cite{greencolor,fairchild2013color} for broader background.

\paragraph{Computational color constancy.} 
A large body of work in \emph{camera calibration} addresses the correction of scene illumination in photos, assuming that perceptual colors remain consistent under different lighting conditions~\cite{gijsenij2011computational, brainard1997bayesian, gehler2008bayesian}. 
This problem is commonly known as \emph{color constancy} or \emph{white balance}. 
Traditional methods rely on hand-crafted assumptions~\cite{buchsbaum1980spatial,finlayson2004shades,van2007edge}, whereas recent learning-based approaches estimate illuminants directly from data~\cite{barron2015convolutional,barron2017fast,bianco2015color,hu2017fc4,afifi2021cross,kim2025ccmnet,afifi2025time}, forming the field of \emph{computational color constancy}. However, due to the high-dimensional nature of light, a global linear $3{\times}3$ transform is insufficient for accurate camera color constancy~\cite{finlayson2014reproduction,cheng2015beyond,karaimer2018improving}. 
Recent work addresses this limitation by either extending single-illuminant estimation to multi-illuminant settings~\cite{kim2021large,afifi2022auto,kim2024attentive,serrano2025revisiting} or moving from RGB alignment to spectral modeling~\cite{koskinen2024single,li2025multi,cogo2025leveraging}. 
These methods improve camera-side color correction, but do not model downstream display reproduction. As a result, they cannot guarantee that the corrected image, once shown on a specific display, will reproduce the perceived appearance of the original scene for a target observer.

\paragraph{Display color management.}
\emph{Display calibration} typically follows ICC workflows, which transform device-dependent signals through a device-independent reference space. 
Chromatic adaptation transforms (e.g., Bradford or von~Kries) compensate for illuminant differences~\cite{hellwig2022brightness}, while color appearance models such as CIECAM02 further account for perceptual viewing conditions~\cite{zhao2007effect,moroney2002ciecam02}. 
Display characterization itself is often modeled using parametric transfer functions, such as Gain–Offset–Gamma (GoG) models~\cite{cho2006inverse,wu2023color}, which describe the mapping between device input signals and emitted radiance. 
Modern devices further adapt these transformations to ambient lighting conditions (e.g., Apple's True Tone~\cite{chang2019ambient,apple2023ambientcolor,qian2024sunlight}). 
These methods characterize display-side reproduction under viewing conditions, but ignore the camera capture process, leaving the coupled capture-to-display problem unaddressed.
\section{Derivation of Color-Accurate Pass-Through}
\label{sec_3-Modeling_Color_Accurate_Pass_Through}

We first introduce a theoretical model for \emph{color pass-through} via a camera $\mathbf{C}$ and a display $\mathbf{D}$, as perceived by a target observer $\mathbf{M}$. Directly shown the captured image on the display may introduce perceptual gap between actual scene color and display color. To mitigate this gap, we apply a correction mapping $\mathbf{F}$ to the captured image before display, counteracting the distortions (\cref{fig_2} left-side).

\subsection{Preliminary: Idealized Radiance Pass-Through}
\label{sec_3.1-Radiance_Pass_Through}

An idealized case of color reproduction is~\emph{radiance pass-through}: for every scene point $i$, the corresponding display radiance $\mathbf{s}^{*}_i$ should precisely reproduce the scene radiance $\mathbf{s}_i$, that is $\mathbf{s}^{*}_i\;=\;\mathbf{s}_i$. This guarantees that \emph{any} observer with spectral sensitivities $\mathbf{M}$ perceives identical colors, as $\mathbf{M}\mathbf{s}^{*}_i=\mathbf{M}\mathbf{s}_i$ holds for $\forall~\mathbf{M}$.

However, this ideal radiance pass-through is in general impossible, as the scene radiance is an high-dimension information, but cameras only record a 3-dimensional color. The coupled camera--display acts as an autoencoder: it encodes a high-dimensional radiance into low-dimensional colors and decodes it back, inevitably discarding spectral information during the transfer process.

One potential solution is to modify the captured image such that the display radiance matches the scene radiance. 
However, this is generally impossible.  We model a camera by its spectral sensitivities $\mathbf{C}\in\mathbb{R}^{3\times L}$ and a display by its spectral primaries $\mathbf{D}\in\mathbb{R}^{L\times 3}$. Let $\mathbf{s}_i,\,\mathbf{s}^{*}_i\in\mathbb{R}^{L}$ denote the discretized scene and reproduced radiance at pixel $i$, sampled at $L$ wavelengths, $3$ be the number of color channels. To achieve radiance pass-through (no gap between scene and display radiance), we need to design a correction function matrix $\mathbf{F}\in\mathbb{R}^{3\times 3}$ that applied to the captured image, such that:
\begin{equation}
\mathbf{s}_i\;\equiv\;\mathbf{s}^{*}_i\;\equiv\;\mathbf{D}\mathbf{F}\mathbf{C}\mathbf{s}_i,
\quad \forall i, 
\label{eq_3.1}
\end{equation} 

For Eq.~\ref{eq_3.1} to hold for all $\mathbf{s}_i$, the composite operator $\mathbf{D}\mathbf{F}\mathbf{C}\in\mathbb{R}^{L\times L}$ must be the identity. However, this is generally impossible: since the correction $\mathbf{F}\in\mathbb{R}^{3\times 3}$, the rank of $\mathbf{D}\mathbf{F}\mathbf{C}$ is at most $3$, and therefore it 
cannot equal the identity in $\mathbb{R}^{L\times L}$ when $L\gg3$. This rank argument formalizes a fundamental limitation: with only finitely many channels, an arbitrary radiance cannot be reconstructed precisely. We therefore target a weaker version of pass-through, discussed below.

\subsection{Color Pass-Through}
\label{sec_3.2-Color_Pass_Through}
While exact \emph{radiance pass-through} is generally impossible, we instead target a more practical goal: \emph{color pass-through}: the display radiance need not match the scene radiance exactly; rather, it should appear similar to an observer (\cref{fig_2} right-side). Here the observer can be either a human or a three-channel camera, both of which possess limited color perceptron. We name this \emph{color pass-through}.

Formally, let $\mathbf{M}\in\mathbb{R}^{3\times L}$ denote the spectral sensitivities of an observer. To achieve color pass-through, the observed display color $\mathbf{M}\mathbf{s}_i\in\mathbb{R}^{3}$, for a given scene point should equal the observed scene color $\mathbf{M}\mathbf{s}^{*}_i\in\mathbb{R}^{3}$. Here we still introduce a correction $\mathbf{F}_{\mathbf{M}}$ on captured image to enforce perceptually equivalence:
\begin{equation}
\mathbf{M}\mathbf{s}_i \;\equiv\; \mathbf{M}\mathbf{s}^{*}_i
\;\equiv\; \mathbf{M}\mathbf{D}\mathbf{F}_{\mathbf{M}}\mathbf{C}\mathbf{s}_i,
\quad \forall i
\quad\Rightarrow\quad
\mathbf{F}_{\mathbf{M}} \;=\;(\mathbf{M}\mathbf{D})^{\dagger}\mathbf{M}\mathbf{C}^{\dagger},
\label{eq_3.2}
\end{equation}
where $(\cdot)^{\dagger}$ denotes the Moore--Penrose pseudoinverse. 
Here $\mathbf{M}\mathbf{D}\in\mathbb{R}^{3\times3}$ and $\mathbf{M}\mathbf{C}^{\dagger}\in\mathbb{R}^{3\times L}$. In practice, directly instantiating the correction $\mathbf{F}_{\mathbf{M}}\in\mathbb{R}^{3\times L}$ is difficult because the sensitivity $\mathbf{M}$ is an unknown high-dimentional matrix and varies across observers. Consequently, $(\mathbf{M}\mathbf{D})^{\dagger}$ cannot be reliably calibrated. 
We therefore propose to approximate $\mathbf{F}_{\mathbf{M}}$ through decomposition described below.

\section{Learning Color Pass-Through}
\label{sec_4}
\begin{figure*}[t] % t/b/h/H
  \centering
  \includegraphics[width=\linewidth]{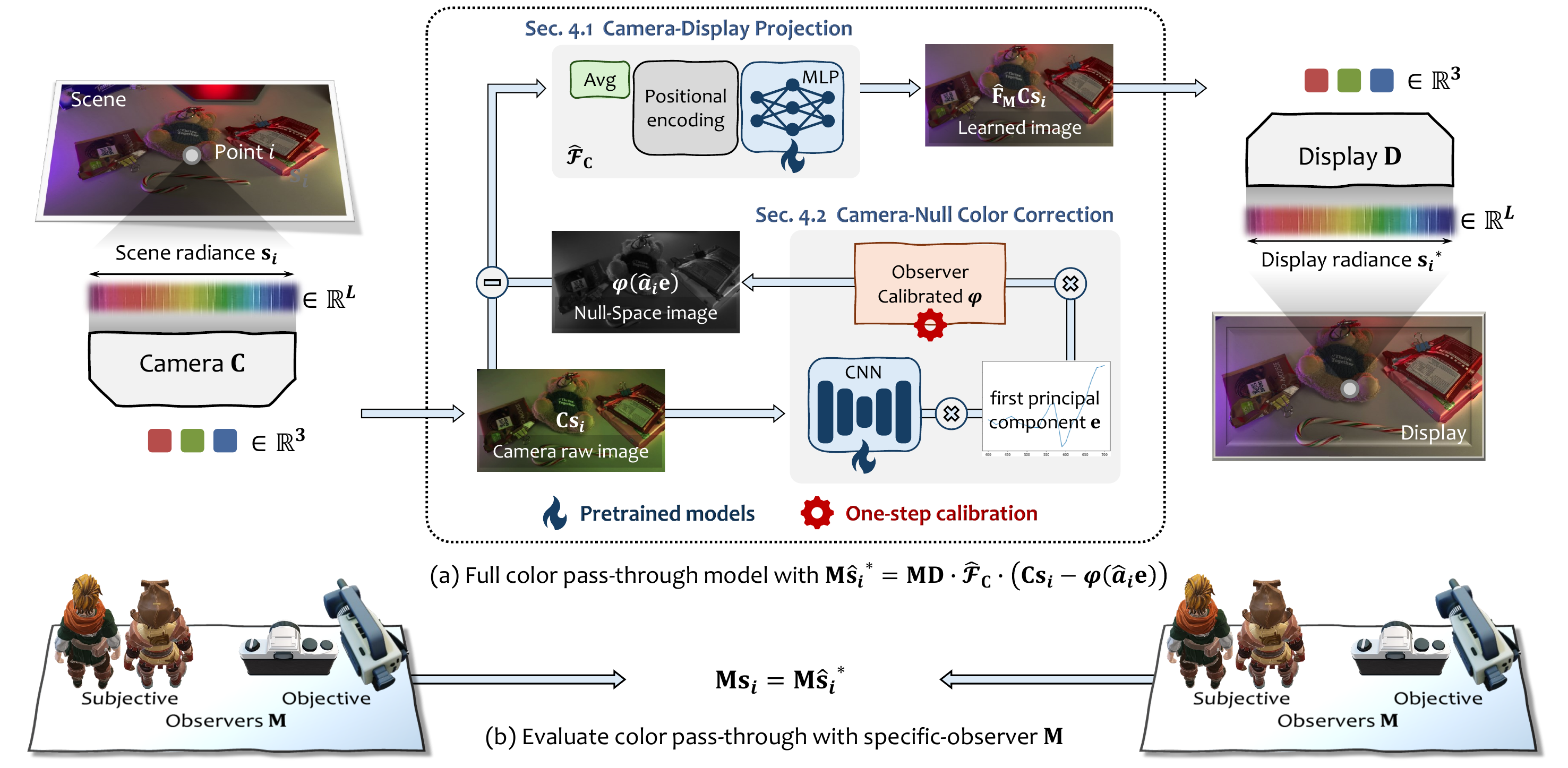} % .pdf/.png/.jpg
  \caption{\textbf{Overall System of Our Color Pass-Through.}
    (a) Full color pass-through model corresponding to~\cref{eq_4.9};
    (b) Objective function and evaluation criteria.}
  \label{fig_3}
\end{figure*}

Since it is hard to directly learn an observer-specifc $\mathbf{F}_{\mathbf{M}}$, we derive a robust, practical solution by decomposing it into two objectives: 
(i) a \emph{camera--display projection} term $\mathcal{F}_{\mathbf{C}}$, which enforces pass-through in the camera measurement space, and 
(ii) a \emph{camera-null color correction} term $\delta_i$, which compensates observer-dependent deviations lies in the spectral space that are invisible to the camera. 

This decomposition turns $\mathbf{F}_{\mathbf{M}}$ into two learnable predictors that can be trained and optimized independently and then combined end-to-end at inference time to calibrate a coefficient $\varphi\in\mathbb{R}^{3\times1}$ within one step for a specific observer, as illustrated in~\cref{fig_3}. The mathematical derivation and implementation details of each module are described below, leading to the final formulation in~\cref{eq_4.9}.

\subsection{Camera-Display Projection}
\label{sec_4.1-Camera_Display_Projection}
In this section, we simplify the problem by replacing the actual human observer $\mathbf{M}$ with the camera $\mathbf{C}$, i.e. $\mathbf{M}:=\mathbf{C}$, where we assume a same-model camera as the observer. Under this condition, \cref{eq_3.2} simplifies to a special case in which the observer-specific objective $\mathbf{F}_{\mathbf{M}}$ collapses to a camera-specific mapping $\mathbf{F}_{\mathbf{C}}$:
\begin{equation}
\mathbf{C}\mathbf{s}_i \;\equiv\; \mathbf{C}\mathbf{s}^{*}_i
\;\equiv\; \mathbf{C}\mathbf{D}\mathbf{F}_{\mathbf{C}}\mathbf{C}\mathbf{s}_i,
\quad \forall i
\quad\Rightarrow\quad
\mathbf{F}_{\mathbf{C}} \;=\; (\mathbf{C}\mathbf{D})^{\dagger},
\label{eq_4.1}
\end{equation}

This yields a simple solution for color pass-through, $\mathbf{F}_{\mathbf{C}}=(\mathbf{C}\mathbf{D})^{\dagger}\in\mathbb{R}^{3\times3}$, which we refer to as the \emph{camera--display projector}. 
It forms a central component of our model and, as we show later, can be learned from data via a simple training objective. The resulting \emph{camera--display projection} is defined as $\mathbf{P}_{\mathrm{C}}:=\mathbf{D}\mathbf{F}_\mathrm{C}\mathbf{C}\in\mathbb{R}^{L\times L}$ as it is an idempotent projection satisfying $\mathbf{P}_{\mathrm{C}}^{2}=\mathbf{P}_{\mathrm{C}}$, which can map any scene radiance $\mathbf{s}_i$ to a \emph{camera metamer} $\mathbf{s}_i^{*}$ that preserves the camera color.

\begin{figure}[t]
\centering
  \includegraphics[width=\linewidth]{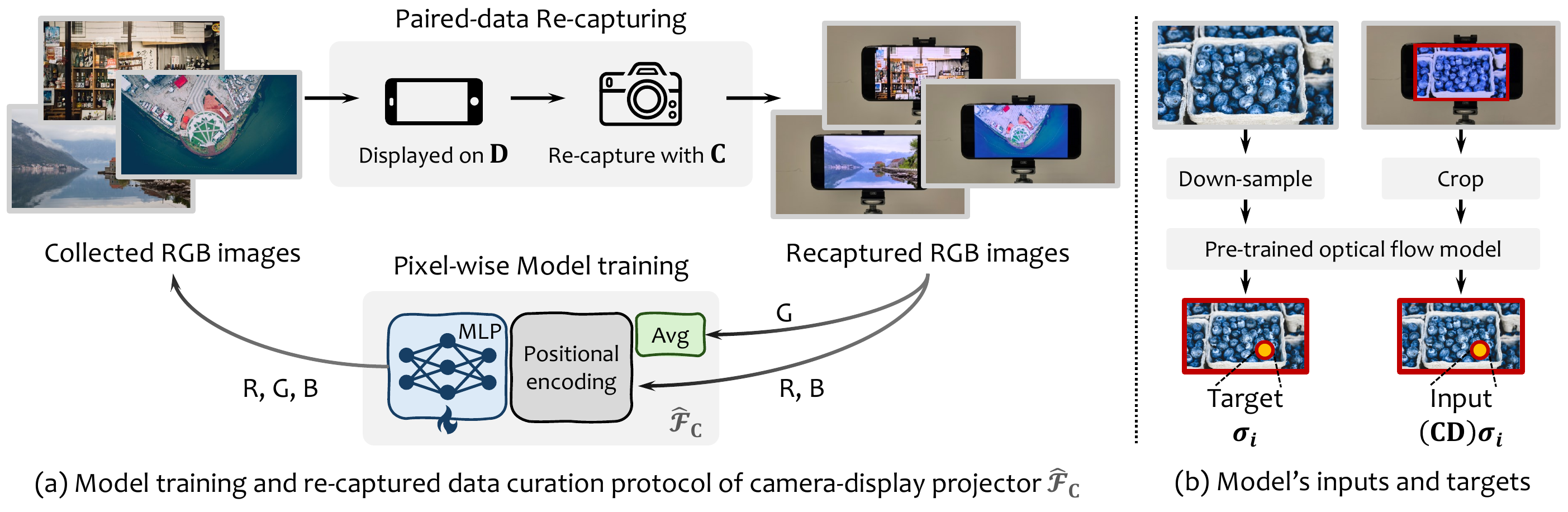}
  \caption{\textbf{Learning $\mathcal{F}_{\mathbf{C}}$ from Recaptured Data.} A pixel-wise model is trained to approximate $\mathcal{F}_{\mathbf{C}}$ via (a) a re-capture protocol and (b) pixel-aligned input–target pairs.}
  \label{fig_4}
\end{figure}

\begin{figure}[t]
\centering
  \includegraphics[width=\linewidth]{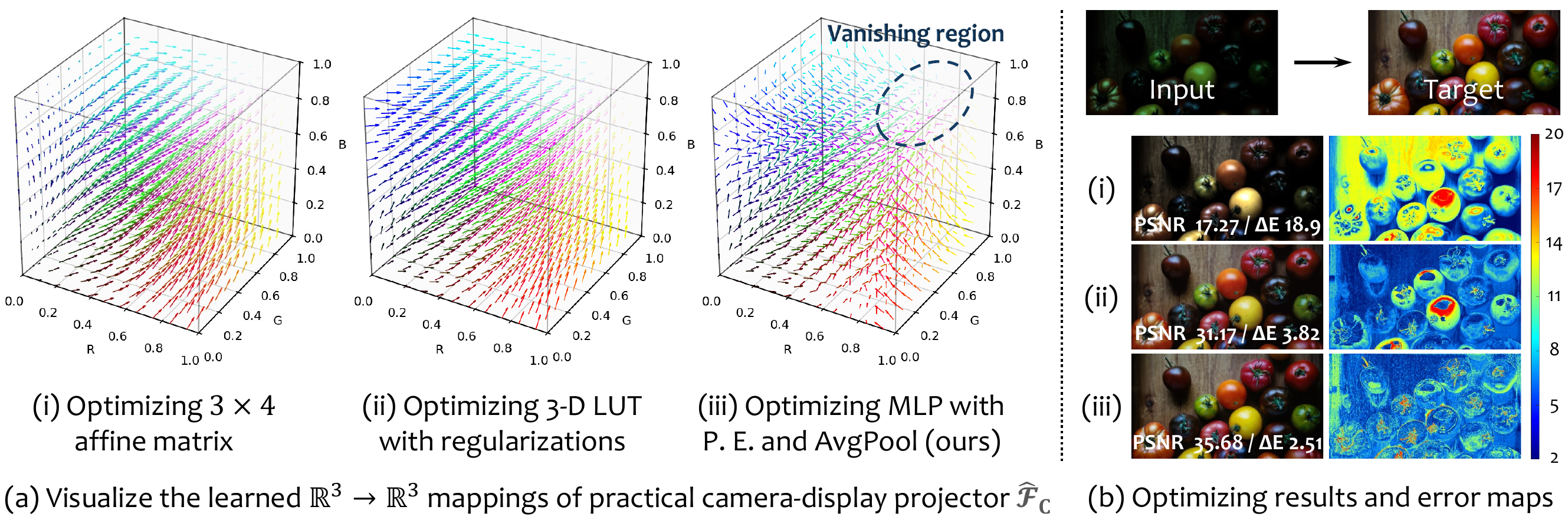}
  \caption{\textbf{Effectiveness of the Learned Model for $\widehat{\mathcal{F}}_{\mathbf{C}}~(\cdot;\theta)$.} In contrast to an affine fit~(i) or a learned 3D LUT~\cite{zeng2020learning}~(ii), our model~(iii): captures (a) higher-frequency details and (b) achieves better performance. Notably, the learned projector exhibits a ``vanishing'' region where multiple camera color collapse to the same displayed color.}
  \label{fig_5}
\end{figure}

In principle, one could estimate $\mathbf{F}_{\mathbf{C}}$ via classical spectral calibration (e.g., we can measure the camera $\mathbf{C}$ with a monochromator and the display $\mathbf{D}$ with a spectrometer). However, such calibration may not be accurate as commercial displays exhibit substantial non-linearities in their default modes (e.g., gamma and tone mapping) that cannot be fully disabled. Consequently, a single affine transform is insufficient to accurately model camera-display projector $\mathbf{F}_{\mathbf{C}}$.

This motivates a data-driven alternative. Instead of enforcing linearity, we treat 
$\mathbf{F}_{\mathbf{C}}=(\mathbf{C}\mathbf{D})^{\dagger}$ as an unknown (potentially non-linear) operator $\mathcal{F}_{\mathbf{C}}~(\cdot)$ and learn it from \emph{re-captured} pairs 
(\cref{fig_4}~(a)). In the forward capture-and-display process, scene radiance passes through the camera 
$\mathbf{C}$ and display $\mathbf{D}$ sequentially, forming the operator $(\mathbf{C}\mathbf{D})$. 
To approximate its inverse, we construct a reverse process: digital images are first rendered on the display 
and then re-captured by the camera, implicitly modeling the pseudo-inverse of $\mathbf{C}\mathbf{D}$.

Specifically, we construct training data by (i) sampling RGB images as supervision $\sigma_i\in\mathbb{R}^3$, (ii) rendering on $\mathbf{D}$, and (iii) re-capturing the displayed images with $\mathbf{C}$ to obtain $(\mathbf{C}\mathbf{D})\sigma_i\in\mathbb{R}^3$. After pixel-wise alignment using an optical-flow model~\cite{teed2020raft} (\cref{fig_4}~(b)), we train a pixel-wise network to explicitly learn the mapping $(\mathbf{C}\mathbf{D})\sigma_i\to\sigma_i$, yielding a non-linear approximation of $(\mathbf{C}\mathbf{D})^{\dagger}$. 

We introduce two key architectural refinements that enable a neural surrogate to represent the practically complex projector $\mathcal{F}_{\mathbf{C}}~(\cdot)$. First, following learned color-transfer models~\cite{le2023gamutmlp,canham2025gain},we use a lightweight multilayer perceptron (MLP) parameterized by $\theta$--two fully connected layers with hidden width 128--augmented with positional encoding to better preserve high-frequency variations. Second, we apply a simple but effective AvgPool layer to the green channel of the input. The motivation is that demosaicing introduces spatial interpolation: each RGB triplet is partially synthesized from neighboring sensor samples. Although pixel-shift cameras could provide per-pixel tri-stimulus measurements~\cite{liu2018subpixelSRcamera}, we instead adopt this minimal preprocessing step, which we find consistently improves accuracy, detailed in supplementary. With these refinements, we adopt a learned neural projector $\widehat{\mathcal{F}}_{\mathbf{C}}(\cdot;\theta)$ and optimize $\theta$ by minimizing the following objective:
\begin{equation}
\theta^{*}
\;=\;\arg\min_{\theta}\sum_{i}
\left\lVert
\,\sigma_i
-
\widehat{\mathcal{F}}_{\mathbf{C}}\!\bigl((\mathbf{C}\mathbf{D})\sigma_i;\theta\bigr)
\right\rVert_{1},
\label{eq_4.2}
\end{equation}

\begin{figure}[t]
\centering
  \includegraphics[width=\linewidth]{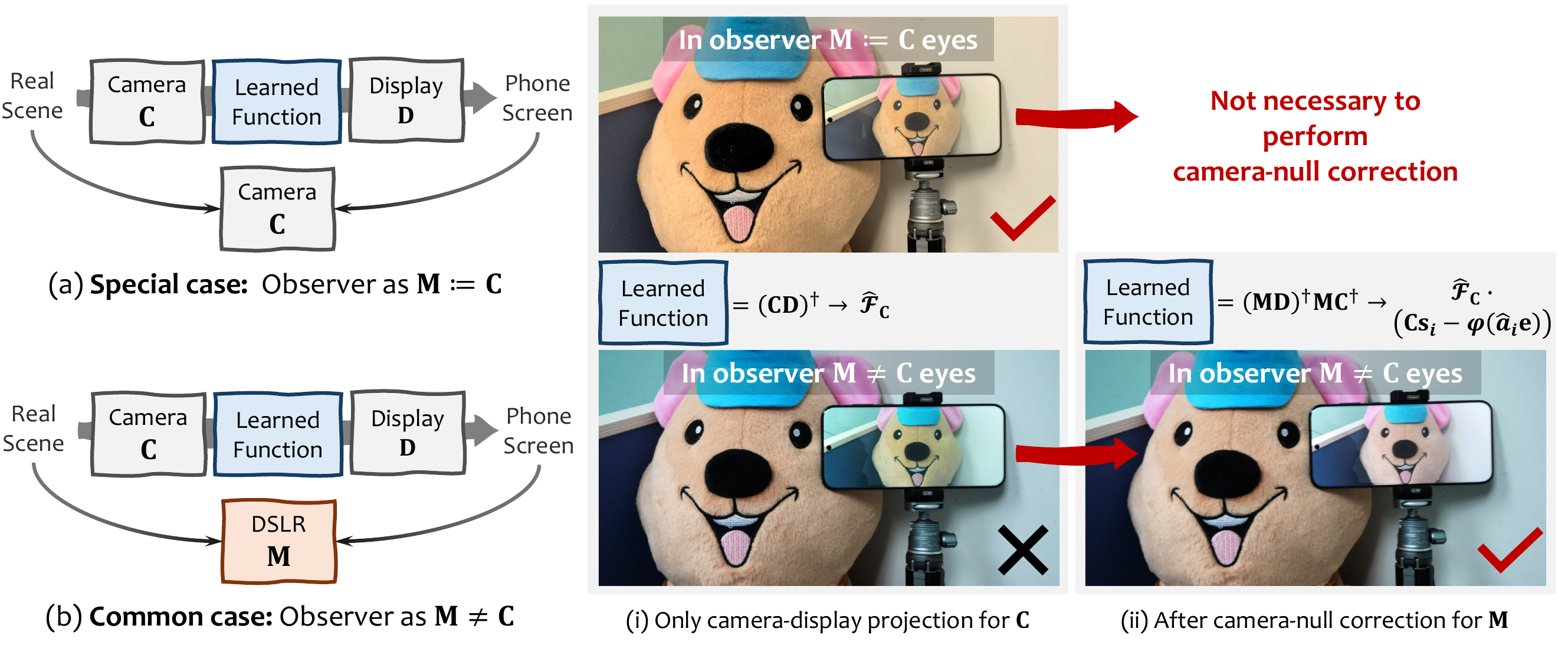}
    \caption{\textbf{Effectiveness of the Two Learned Components for Color Pass-Through.}
    (a) When using another camera of the same model as the digital observer: 
    (i) the learned camera--display projection alone is sufficient to achieve pass-through, 
    (ii) no observer calibration for $\mathbf{M}$ is required; 
    (b) When observer is different from camera (e.g. a DSLR): 
    (i) the camera--display projection alone is insufficient, 
    (ii) adding the camera-null correction for a target observer $\mathbf{M}$ yields consistent color reproduction.}
  \label{fig_6}
\end{figure}

Compared to alternative learned fits, our model captures substantially higher-frequency details (\cref{fig_5}~(a)) and yields better quantitative results (\cref{fig_5}~(b)). However, the learned camera--display projector guarantees color pass-through only when camera itself is the observer (\cref{fig_6}~(a)).When the observer differs from the camera---e.g., a DSLR---$\widehat{\mathcal{F}}_{\mathbf{C}}$ alone leads to noticeable color shifts in the image displayed on the phone (\cref{fig_6}~(b)), we address this in the next section.

\subsection{Camera-Null Color Correction}
While the learned projector $\widehat{\mathcal{F}}_{\mathbf{C}}$ (approximating $(\mathbf{C}\mathbf{D})^{\dagger}$) reliably enforces color pass-through \emph{in the camera measurement space}, our ultimate goal is pass-through for a target observer $\mathbf{M}$. We therefore analyze the conditions under which the camera-aligned projector transfers to other observers and when it fails.

Our motivation is empirical. Consider two smartphones of the same model: one acts as the camera and the other as the observer (\cref{fig_6}~(a)). When we compare the real scene to the image displayed on phone through an identical phone (a digital observer), the camera--display projector alone reproduces matching colors from phone's viewpoint. Yet the color of the displayed image on phone can still deviate from the real scene perceptually for other viewers. In practice, we often observe a faint tinted ``color mask'' over the screen. To make this discrepancy explicit, we introduce a DSLR as a proxy 3-channel observer (\cref{fig_6}~(b)).

To eliminate this ``color mask'', we introduce \emph{camera-null color correction}. We estimate the residual color cast using a single observer-specific calibration coefficient and compensate for it to achieve color pass-through for $\mathbf{M}$.

First, we derive the origin of the residual color cast. 
For clarity of the derivation, we temporarily treat $\mathbf{F}_{\mathbf{C}}$ as linear and apply the correction at the \emph{input} to the projector rather than at its output. 
The reason is practical: in real systems $\mathbf{F}_{\mathbf{C}}$ is intrinsically non-linear (see~\cref{fig_5}), making compensation \emph{after} this mapping more entangled and less stable. 
Applying the correction \emph{before} the projector instead yields a cleaner and more robust adjustment. Therefore, we introduce a correction term $\delta_i\in\mathbb{R}^{3}$ at the input of $\mathbf{F}_{\mathbf{C}}$ and relate the observer-specific solution in~\cref{eq_3.2} to the camera-specific solution in~\cref{eq_4.1}, yielding:
\begin{equation}
\mathbf{F}_{\mathbf{M}}\,\mathbf{C}\mathbf{s}_i
\;=\;
\mathbf{F}_{\mathbf{C}}\bigl(\mathbf{C}\mathbf{s}_i-\delta_i\bigr)
\quad\Rightarrow\quad
\delta_i
\;:=\;
\mathbf{C}\bigl(\mathbf{I}-\mathbf{P}_{\mathbf{M}}\bigr)\mathbf{s}_i,
\label{eq_4.3}
\end{equation}
where $\mathbf{P}_{\mathbf{M}}=\mathbf{D}(\mathbf{M}\mathbf{D})^{\dagger}\mathbf{M}$ is an idempotent projection sharing similar properties with $\mathbf{P}_{\mathrm{C}}$. 
\cref{eq_4.3} reveals two key regimes characterizing the correction term $\delta_i$:

\paragraph{(1) when $\delta_i=0$ for all observers $\mathbf{M}$.} 
Sufficient condition for~\cref{eq_4.3} to hold is: 
% (detailed proofs are provided in the supplementary PDF)
\begin{equation}
\mathrm{null}(\mathbf{C}) \subseteq \mathrm{null}(\mathbf{M}),
\label{eq_4.4}
\end{equation}
which states that any spectrum invisible to the camera is also invisible to the observer. 
We refer to this as the \emph{Luther--Ives condition under camera--display projection}. 
It is closely related in spirit to the classical Luther--Ives condition for colorimetric capture 
($\mathbf{M}=\mathbf{T}\mathbf{C} \Leftrightarrow \mathrm{null}(\mathbf{M})=\mathrm{null}(\mathbf{C})$), but extends its condition. We provide the proof in the supplementary and suggest a hardware-driven solution that enforce $\delta_i=0$. However,~\cref{eq_4.4} generally does not hold in common settings. We therefore seek an optimization-based learning method to estimate an observer-specific approximation of $\delta_i$, leading to the second regime.

\paragraph{(2) when $\delta_i\neq 0$ for a specific observer $\mathbf{M}$.}
According to~\cref{eq_4.3}, using $\mathbf{F}_{\mathbf{C}}$ requires estimating $\delta_i$ and subtracting it from the camera color $\mathbf{C}\mathbf{s}_i$. To analyze $\delta_i$ explicitly, we propose to decompose $\mathbf{s}_i$ by project it with projector $\mathbf{P}_{\mathbf{C}}$:
\begin{equation}
\mathbf{s}_i=\mathbf{r}_i+\mathbf{n}_i,\qquad
\left\{
\begin{aligned}
\mathbf{r}_i &:= \mathbf{P}_{\mathrm{C}}\mathbf{s}_i \quad \in \mathrm{Range}(\mathbf{D}),\\
\mathbf{n}_i &:= \mathbf{s}_i-\mathbf{r}_i \quad\in \mathrm{Null}(\mathbf{C}),
\end{aligned}
\right.
\label{eq_4.5}
\end{equation}

Geometrically,~\cref{eq_4.5} corresponds to an oblique projection of $\mathbf{s}_i$ onto $\mathrm{range}(\mathbf{D})$ along $\mathrm{null}(\mathbf{C})$. A key consequence is $\mathbf{C}(\mathbf{I}-\mathbf{P}_{\mathbf{M}})\mathbf{r}_i=\mathbf{0}$, see the supplementary for a proof.
Therefore, the correction term $\delta_i$ of~\cref{eq_4.3} simplifies to $\mathbf{C}(\mathbf{I}-\mathbf{P}_{\mathbf{M}})\mathbf{n}_i$, which shows that the correction term originates entirely from the \emph{metameric black} (i.e., invisible) space to the camera $\mathbf{C}$, since $\mathbf{n}_i\in\mathrm{null}(\mathbf{C})$ and thus $\mathbf{C}\mathbf{n}_i=\mathbf{0}$.

In practice, the only available measurement is the camera color $\mathbf{C}\mathbf{s}_i\in\mathbb{R}^{3}$, yet the desired correction $\delta_i$ depends on the unobserved component $\mathbf{n}_i$. 
A naïve solution would be to treat $\mathbf{C}\mathbf{s}_i$ as a prior and attempt to reconstruct $\mathbf{n}_i\in\mathbb{R}^{L}$ (or even the full spectrum). 
Even if such reconstruction were feasible, a second bottleneck remains: computing $\delta_i$ requires the observer-related operator $\mathbf{C}(\mathbf{I}-\mathbf{P}_{\mathbf{M}})\in\mathbb{R}^{3\times L}$, which is impractical to calibrate since it has $L$ dimensionality.

\begin{figure}[t]
\centering
  \includegraphics[width=0.9\linewidth]{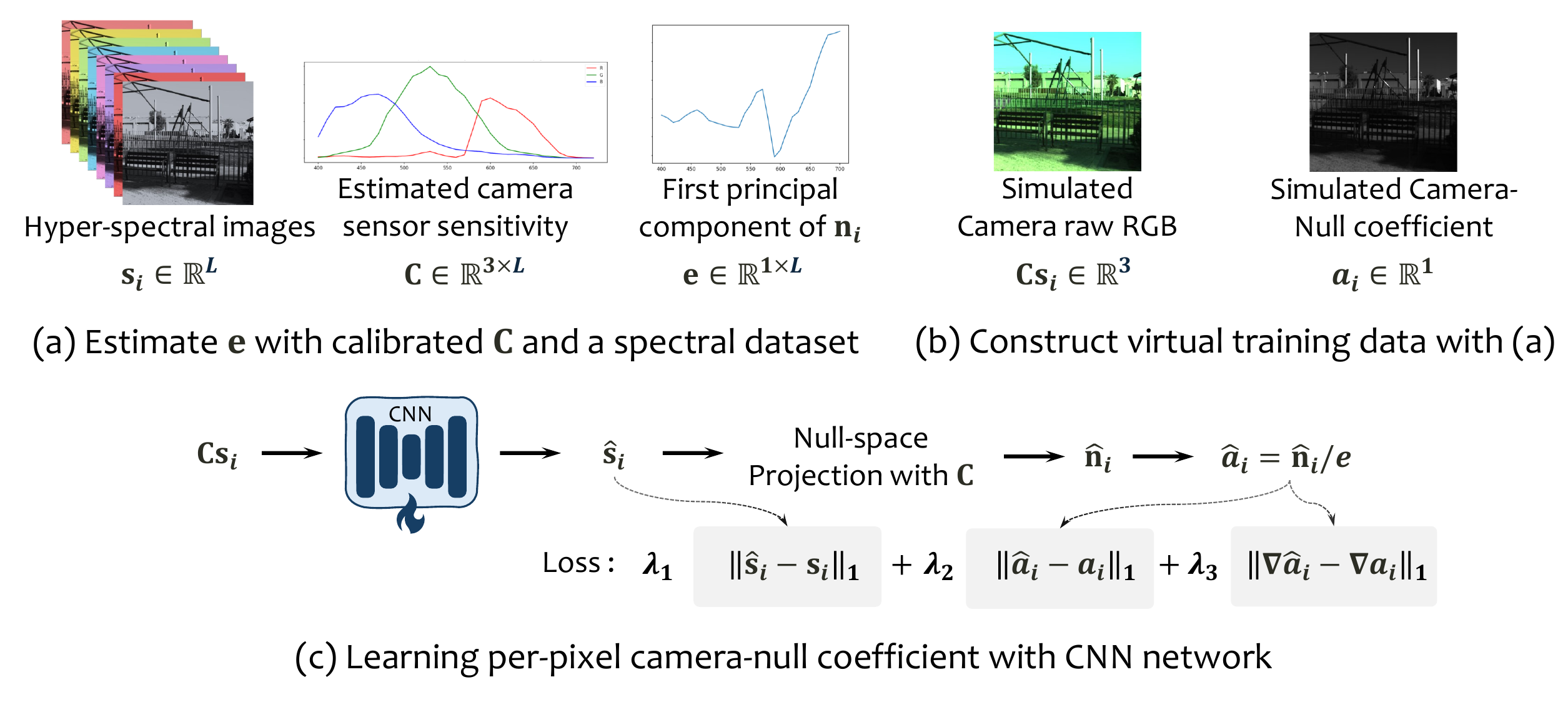}
  \caption{\textbf{Learning Camera-null Color Coefficient ${a}_i$ from Hyper-spectral Data.}}
  \label{fig_4.4}
\end{figure}

To make $\mathbf{C}(\mathbf{I}-\mathbf{P}_{\mathbf{M}})$ calibratable, we use the classic observations that natural reflectance and radiance spectra exhibit low intrinsic dimensionality~\cite{maloney1986evaluation,parkkinen1989characteristic}. We therefore assume that the camera-null component $\mathbf{n}_i$ also lies approximately in a low-dimensional subspace and model it with a PCA basis, we get:
\begin{equation}
\mathbf{n}_i \approx \sum_{k=1}^{K} a_{i}^{(k)}\,\mathbf{e}^{(k)} \qquad K \ll L,
\label{eq_4.6}
\end{equation}
where $K$ is the number of PCA bases. We find that the first principal component explains approximately $93\%$ of the variance on a large radiance set~(see supplement)), motivating our practical rank-1 approximation with $K=1$: 
$\mathbf{n}_i \approx \hat{a}_i\mathbf{e}\in\mathbb{R}^{1\times L}$ and consequently the correction color $\delta_i$ from~\cref{eq_4.3} becomes:
\begin{equation}
\delta_i =\mathbf{C}(\mathbf{I}-\mathbf{P}_{\mathbf{M}})\cdot\mathbf{n}_i \approx \varphi\,\cdot({a}_i\mathbf{e}),
\label{eq_4.7}
\end{equation}
Here, all three terms can be estimated. The first principle component $\mathbf{e}$ can be pre-computed from data; the scalar ${a}_i$ is a \emph{camera-null color coefficient} at pixel $i$ can be estimated via a learned predictor, and $\varphi$ is a \emph{calibration coefficient} only relates to observer $\mathbf{M}$ and display $\mathbf{D}$. This parameterization makes calibration practical, since $\varphi\in\mathbb{R}^{3\times 1}$ contains only 3 unknown variables. Moreover, later in~\cref{fig_13}, we show $\varphi$ remains nearly stable across varying in-the-wild scenes.

Specifically, to learn a predictor for $a_i$, we proceed in three stages, as shown in~\cref{fig_4.4}: (a) we estimate the camera spectral sensitivities $\mathbf{C}$ via a ColorChecker calibration procedure, following the robust method of~\cite{jiang2013space}; using the calibrated $\mathbf{C}$, we then compute a PCA basis for the camera-null component from hyperspectral datasets (treated as $\mathbf{s}_i$) and retain the dominant component as $\mathbf{e}$; (b) we construct training inputs as simulated camera raw measurements $\mathbf{C}\mathbf{s}_i$, with the learning target being $a_i$; and (c) we estimate the per-pixel coefficient $\hat{a}_i$ with an efficient spectral reconstruction network~\cite{cai2022mst++} by minimizing the loss function:
\begin{equation}
\mathcal{L}
=
\sum_{i}
\Big[
\lambda_1\left\lVert \hat{\mathbf{s}}_i-\mathbf{s}_i \right\rVert_{1}
+
\lambda_2\left\lVert\hat{a}_i-a_i\right\rVert_{1}
+
\lambda_3
\left\lVert\nabla\hat{a}_i-\nabla a_i\right\rVert_{1}
\Big].
\label{eq_4.8}
\end{equation}
where $\nabla$ denotes the total-variation operator applied to the coefficient $a_i$, encouraging spatial smoothness. 
In addition to supervising $\hat{a}_i$ directly, we also retain the spectral reconstruction loss $\lVert \hat{\mathbf{s}}_i-\mathbf{s}_i\rVert_1$. 
Later in~\cref{tab_5.1.2}, we show spectral reconstruction supervision also improves the estimated coefficient $\hat{a}_i$ accuracy.

In summary, combining the observer-specific color pass-through correction of \cref{eq_3.2} with 
the learned camera--display projector $\widehat{\mathcal{F}}_{\mathbf{C}}$ (\cref{eq_4.3}) and the learned camera-null correction $\widehat{\delta}_i=\hat{a}_i\mathbf{e}$ (\cref{eq_4.7}), 
our full Color Pass-Through model is:
\begin{equation}
\underbrace{\mathbf{M}\mathbf{s}_i}_{\text{scene color}}
\approx
\underbrace{\mathbf{M}\widehat{\mathbf{s}}^{*}_i}_{\text{display color}}
=\;
\mathbf{M}\mathbf{D}\;\cdot
\underbrace{\widehat{\mathcal{F}}_{\mathbf{C}}}_{\textbf{learned}}\cdot\;
\bigl(\underbrace{\mathbf{C}\mathbf{s}_i}_{\text{camera color}}
- \underbrace{\varphi}_{\textbf{calibrate}} \cdot
\underbrace{(\hat{a}_i\mathbf{e})}_{\textbf{learned}}\bigr)
% \;=\;
% \underbrace{\mathbf{M}\mathbf{s}_i}_{\text{scene color}}.
\label{eq_4.9}
\end{equation}

After pre-training the two predictors $\widehat{\mathcal{F}}_{\mathbf{C}}$ and $\hat{a}_i\mathbf{e}$ (see~\cref{fig_4} and~\cref{fig_4.4}), we perform a one-time calibration for a target observer to estimate $\varphi\in\mathbb{R}^{3\times1}$. Once calibrated, $\varphi$ is kept fixed for all subsequent evaluations.
\section{RESULTS}
\label{sec_5}
We first evaluate the fitting quality of the two learned predictors, then compare the full color pass-through model against several baselines, and finally present ablations to validate our design choices and method's robustness. Additional experiments are reported in the supplementary material.

All experiments are conducted using two identical smartphones (HUAWEI Pura~70 Pro or Xiaomi~17 Pro Max): one acts as the camera $\mathbf{C}$ and the other as the display $\mathbf{D}$; a DSLR camera (Sony ILCE-7M4) is used as the digital observer $\mathbf{M}$ to produce all quantitative results and figures. 
We set the phone screen to maximum brightness to maximize the usable color gamut and capture linear RAW images in the phone's Pro mode to minimize extra in-camera processing. For indoor multi-illuminant tests, we use Ulanzi VL49RGB lights in HSI mode, whose dedicated RGB LEDs provide spectrally peaky illumination.

\subsection{Fitting Quality on Two Learned Components}
\label{sec_5.1}

\begin{table*}[t]
\centering

\begin{subtable}[t]{0.64\textwidth}
\centering
\caption{Evaluation of Learned Camera-Display Projector $\widehat{\mathcal{F}}_{\mathbf{C}}$.}
\label{tab_5.1.1}
{\footnotesize
\setlength{\tabcolsep}{3.5pt}
\renewcommand{\arraystretch}{1.05}
\resizebox{\textwidth}{!}{%
\begin{tabular}{cc|cc|cccc}
\toprule
\multicolumn{2}{c}{\textbf{Methods}} & Params (K) & Runtime (ms) &
PSNR $\uparrow$ & $\Delta E_{\text{mean}}\downarrow$ & $\Delta E_{p95}\downarrow$ & STRESS $\downarrow$ \\
\midrule
\multicolumn{2}{c}{IA-3DLUT}~\cite{zeng2020learning} & 593.0 & 339.92 &
26.69 & \third{5.66} & \second{9.22} & 9.22 \\
\multicolumn{2}{c}{NiLUT}~\cite{conde2024nilut} & 33.9 & 31.16 &
\third{31.03} & \second{3.59} & 9.39 & \third{5.02} \\
\multicolumn{2}{c}{CSRNet}~\cite{he2020conditional} & 36.5 & 253.78 &
\second{31.12} & \second{3.59} & \third{9.34} & \second{5.00} \\
\multicolumn{2}{c}{Ours} & 30.8 & 41.30 &
\best{32.13} & \best{3.34} & \best{8.81} & \best{4.69} \\
\bottomrule
\end{tabular}}
}
\end{subtable}
\hfill
\begin{subtable}[t]{0.35\textwidth}
\centering
\caption{Evaluation of Learned $\hat{a}_i$}
\label{tab_5.1.2}
{\footnotesize
\setlength{\tabcolsep}{4.5pt}
\renewcommand{\arraystretch}{1.0}
\resizebox{\textwidth}{!}{%
\begin{tabular}{ccc|cc}
\toprule
$\lambda_1$ & $\lambda_2$ & $\lambda_3$ & PSNR($\hat a_i, a_i$)$\uparrow$ & PSNR($\hat{\mathbf{s}}_i,\mathbf{s}_i$)$\uparrow$\\
\midrule
\checkmark & & & 28.27 & 14.94 \\
& \checkmark & \checkmark & \second{29.42} & \best{32.69} \\
\checkmark & \checkmark & & \third{29.00} & \third{31.49} \\
\checkmark & \checkmark & \checkmark & \best{29.68} & \second{31.95} \\
\bottomrule
\end{tabular}}
}
\end{subtable}

\caption{Evaluate performance of learned models for camera–display projector $\widehat{\mathcal{F}}_{\mathbf{C}}$ and camera–null color coefficients $\hat{a}_i$. We report quantitative metrics, including PSNR~\cite{gonzalez2008digital}, $\Delta E_{mean}$~\cite{sharma2005ciede2000}, $\Delta E*{95}$ (95th percentile), and STRESS~\cite{finlayson2015reproduction}, as well as runtime on 2K-resolution inputs measured on an NVIDIA RTX~4090 GPU.}
\label{tab_5.1}

\end{table*}

\begin{figure*}[t] % t/b/h/H
  \centering
  \includegraphics[width=\linewidth]{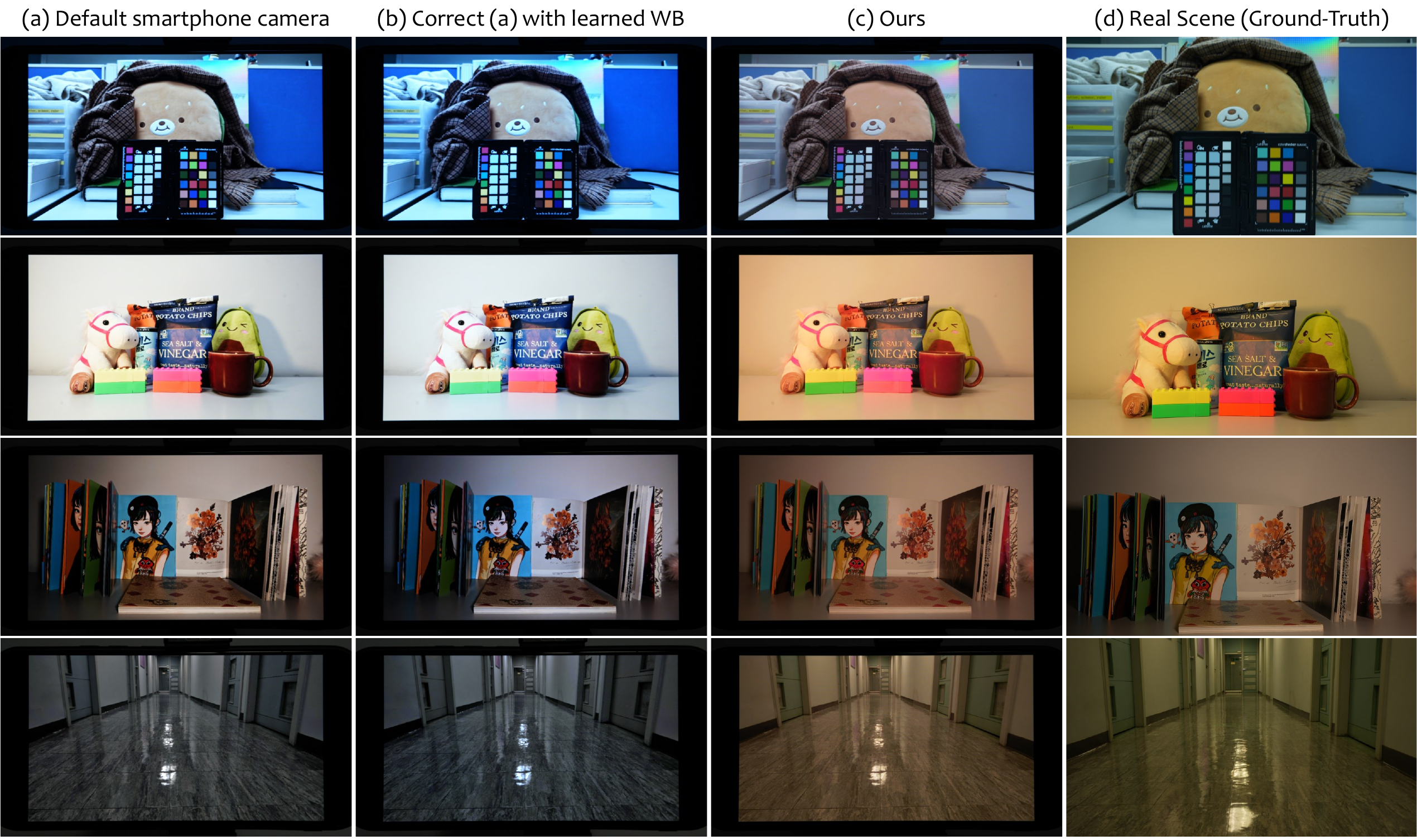} % .pdf/.png/.jpg
  \caption{\textbf{Separated-view Comparison.} The left three columns show images observed through a phone display, while the rightmost column directly observes the real scene. Our method preserves more consistent colors than other baselines.}
  \label{fig_8}
\end{figure*}

\begin{figure*}[t] % t/b/h/H
  \centering
  \includegraphics[width=\linewidth]{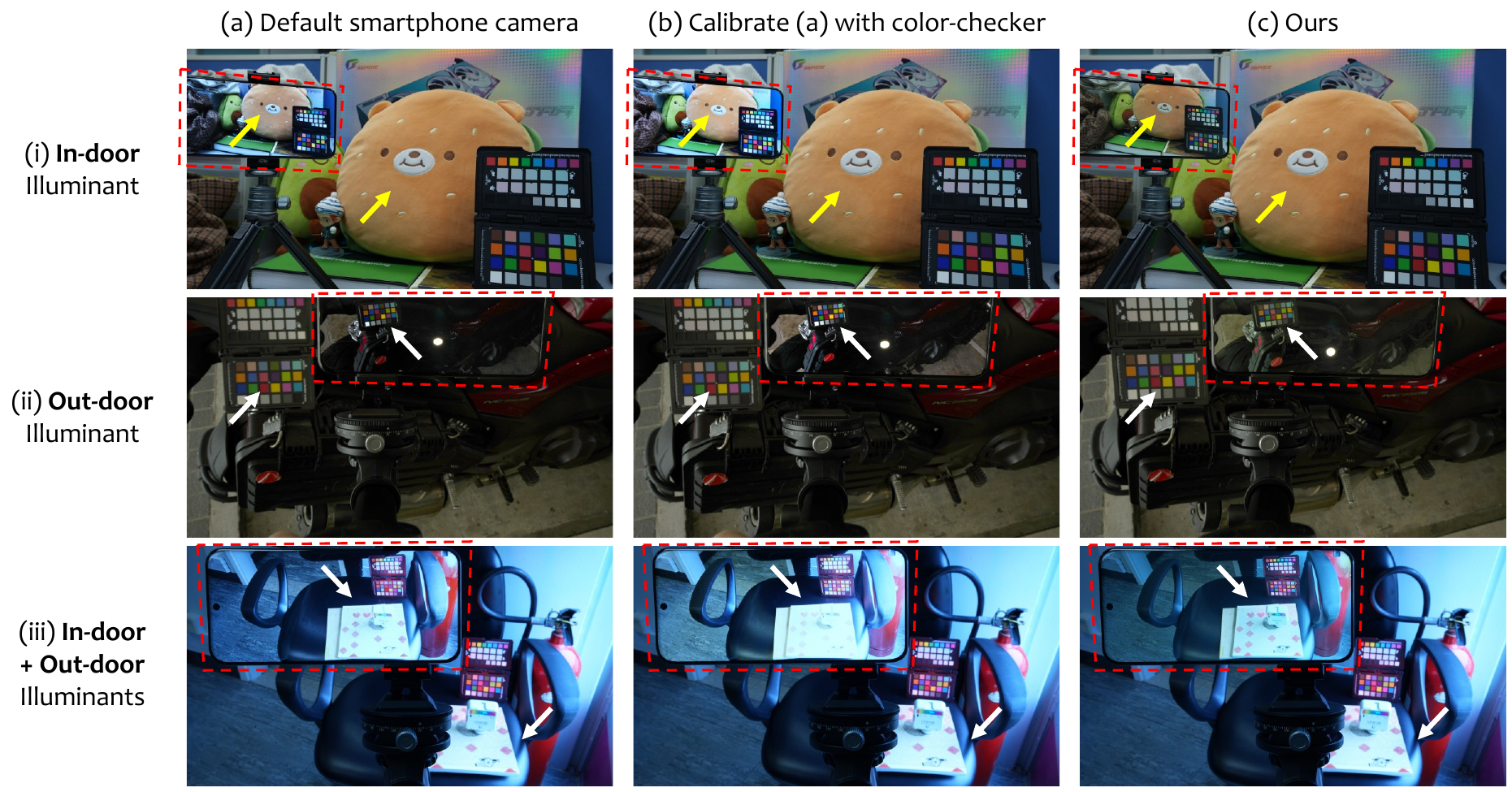} % .pdf/.png/.jpg
  \caption{\textbf{Direct in-scene Comparison.} Each column shows a smartphone screen with the processed image hovering over the real scene. Our method reproduces closer colors. }
  \label{fig_9}
\end{figure*}

\subsubsection{Evaluation on Learned Camera-Display Projector $\widehat{\mathcal{F}}_{\mathbf{C}}$.}
\label{sec_5.1.1}
We train $\widehat{\mathcal{F}}_{\mathbf{C}}$ using the DIV2K dataset~\cite{Agustsson_2017_CVPR_Workshops}, denoised by~\cite{li2024dualdn}, and perform recapture process to obtain 800 pairs for training and 100 for testing. We report evaluation in~\cref{tab_5.1.1} and compare our learned model with existing pixel-wise color transfer methods.

\subsubsection{Evaluation on Learned Camera-Null Color Coefficient $\hat{a}_i$.}
\label{sec_5.1.2}
We train $\hat{a}_i$ using hyper-spectral datasets (ARAD-1K~\cite{arad2022ntire}, CAVE~\cite{yasuma2010generalized} and ICVL~\cite{arad2016sparse}) with $L=31$ spectral channels. In total, we use 1000 images for training and 182 for testing. We report evaluation in~\cref{tab_5.1.2} under different loss configurations. Based on the setting that achieves the highest accuracy for $\hat{a}_i$, we preserve $\lambda_1,\lambda_2$ and adopt $\lambda_1,\lambda_2,\lambda_3=(1,0.01,1)$ in~\cref{eq_4.8} for all subsequent experiments.

\begin{table}[t]
\centering
\caption{Quantitative comparisons for color pass-through using the 24 ColorChecker patches under unseen, diverse illumination. Numbers in gray indicate results with brightness aligned to our method; our method still achieves the best performance.}
\label{tab_5.2}
\scriptsize
\setlength{\tabcolsep}{2pt}
\renewcommand{\arraystretch}{1}
\resizebox{\linewidth}{!}{%
\begin{tabular}{cccccccc}
\toprule
\multicolumn{2}{l}{\textbf{Methods}} &
\multicolumn{2}{c}{PSNR $\uparrow$} &
\multicolumn{2}{c}{$\Delta E_{\text{mean}}\downarrow$} &
\multicolumn{2}{c}{STRESS $\downarrow$} \\
\cmidrule(lr){3-4}\cmidrule(lr){5-6}\cmidrule(lr){7-8}
& & Huawei & Xiaomi & Huawei & Xiaomi & Huawei & Xiaomi \\
\midrule
\multicolumn{2}{l}{Default smartphone camera} &
13.78 \extra{15.80} & 14.61 \extra{15.93} &
\third{14.62} \extra{12.31} & \third{13.49} \extra{11.54} &
\third{26.23} \extra{27.58} & \third{25.07} \extra{25.27} \\
\multicolumn{2}{l}{Color-checker calibration} &
\third{15.02} \extra{15.23} & \third{16.36} \extra{16.85}&
18.49 \extra{18.40} & 15.32 \extra{15.09} &
38.85 \extra{38.56} & 36.42 \extra{36.04} \\
\multicolumn{2}{l}{Multi-illuminants Auto-WB} &
12.84 \extra{12.95} & 13.92 \extra{14.41} &
17.84 \extra{17.37} & 17.08 \extra{16.26} &
31.12 \extra{30.48} & 30.05 \extra{29.66} \\
\multicolumn{2}{l}{Ours w/o camera-null correction} &
\second{27.32} & \second{27.84} &
\second{6.49} & \second{5.97} &
\best{17.27} & \second{16.39} \\
\multicolumn{2}{l}{Ours} &
\best{28.65} & \best{29.10} &
\best{5.18} & \best{4.79} &
\second{17.48} & \best{16.12} \\
\bottomrule
\end{tabular}
}
\end{table}

\subsection{Evaluating Full Color Pass-Through Model}
\label{sec_5.2}

\subsubsection{Compared Methods.}
We compare against three strong and practical baselines that cover the standard camera-side and ICC-style correction pipelines.
(a) \emph{Default smartphone camera:} the image produced by the smartphone's commercial ISP, including its proprietary white balance, tone mapping, and color tuning. This represents a highly optimized commercial pipeline.
(b) \emph{Multi-illuminant Auto-WB:} a state-of-the-art learned white-balance method~\cite{afifi2022awb} applied to the default smartphone output to correct spatially varying illumination-induced color casts.
(c) \emph{ColorChecker calibration:} an instrumented calibration baseline using a ColorChecker Passport~\cite{sunoj2018colorcalibration}. We estimate an ICC/profile-based color correction from the ColorChecker measurements and render the corrected image in Photoshop before display, making this a strong calibration baseline and an approximate upper bound for standard per-scene color correction workflows.

\subsubsection{Objective Evaluation.}
We first calibrate the digital observer (DSLR) via a one-time grid search for the calibration coefficient $\varphi\in\mathbb{R}^{3\times1}$ using 24 ColorChecker patches under natural illumination. Each entry of $\varphi$ is searched in $[0.025,0.075]$ with step $0.0125$, and the $\varphi$ that minimizes the error is selected.

We then evaluate our method and all baselines using the same ColorChecker under a diverse set of illuminations: ten correlated color temperatures ranging from $2500\,\mathrm{K}$ to $9000\,\mathrm{K}$, and five randomly sampled RGB-LED illuminant colors, quantitative results over the 24 patches are summarized in~\cref{tab_5.2}. Moreover, qualitative comparisons under diverse real scenes are shown in~\cref{fig_8} and~\cref{fig_9}.

\subsubsection{Subjective Evaluation.}
\begin{figure*}[t] % t/b/h/H
  \centering
  \includegraphics[width=\linewidth]{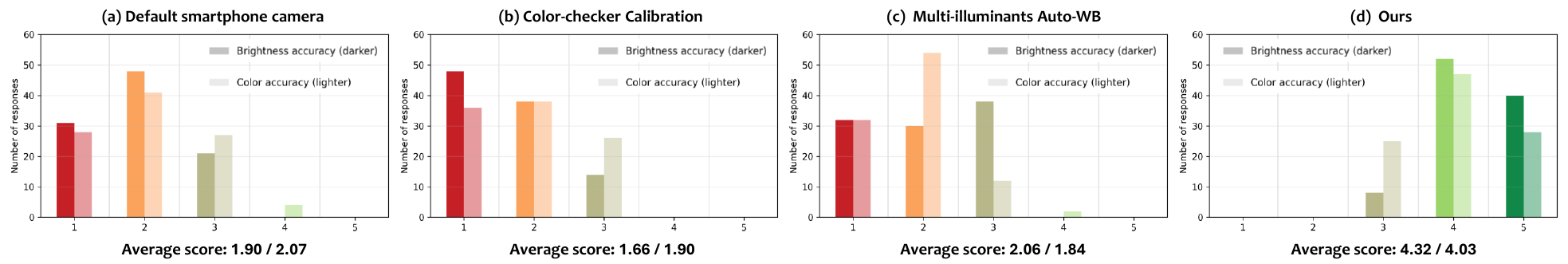} % .pdf/.png/.jpg
  \caption{\textbf{User study on brightness and color accuracy.} Participants rated similarity to the reference on a 5-point Likert scale (1 for least similar, 5 for most similar).}
  \label{fig_10}
\end{figure*}

We conduct user studies with ten human observers to assess perceptual color pass-through. A one-time grid search generates candidate calibration coefficients $\varphi\in\mathbb{R}^{3\times1}$, from which each participant selects the best $\varphi$ in one scene. After calibration, participants evaluate each method across ten unseen scenes by rating \emph{brightness accuracy} and \emph{color accuracy}, measuring how well the displayed image matches the real scene, as summarized in~\cref{fig_10}.

\begin{figure}[t] % t/b/h/H
  \centering
  \includegraphics[width=\linewidth]{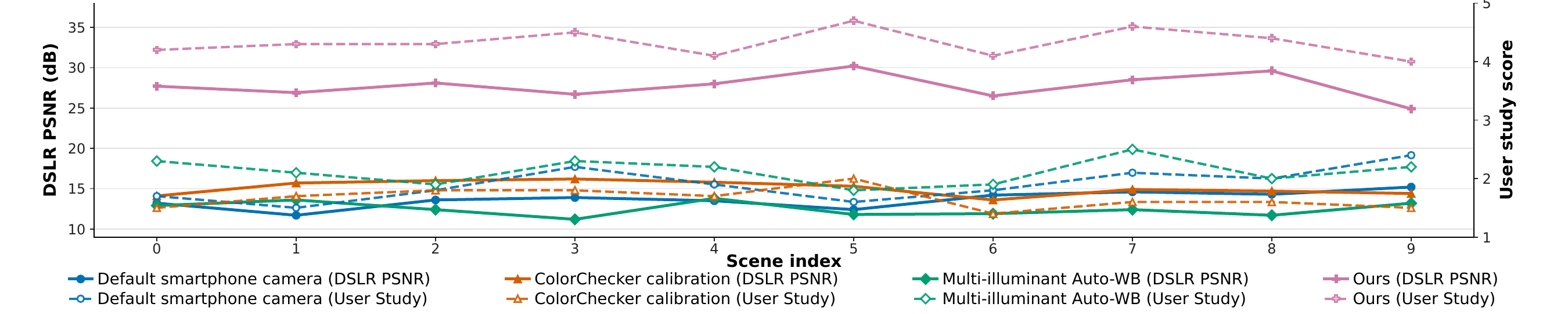} % .pdf/.png/.jpg
  \caption{Preference comparison between DSLR-based and human-subject evaluations.}
  \label{fig_11}
\end{figure}

\begin{figure}[t]
\centering
\begin{minipage}{0.45\linewidth}
    \centering
    \includegraphics[width=\linewidth]{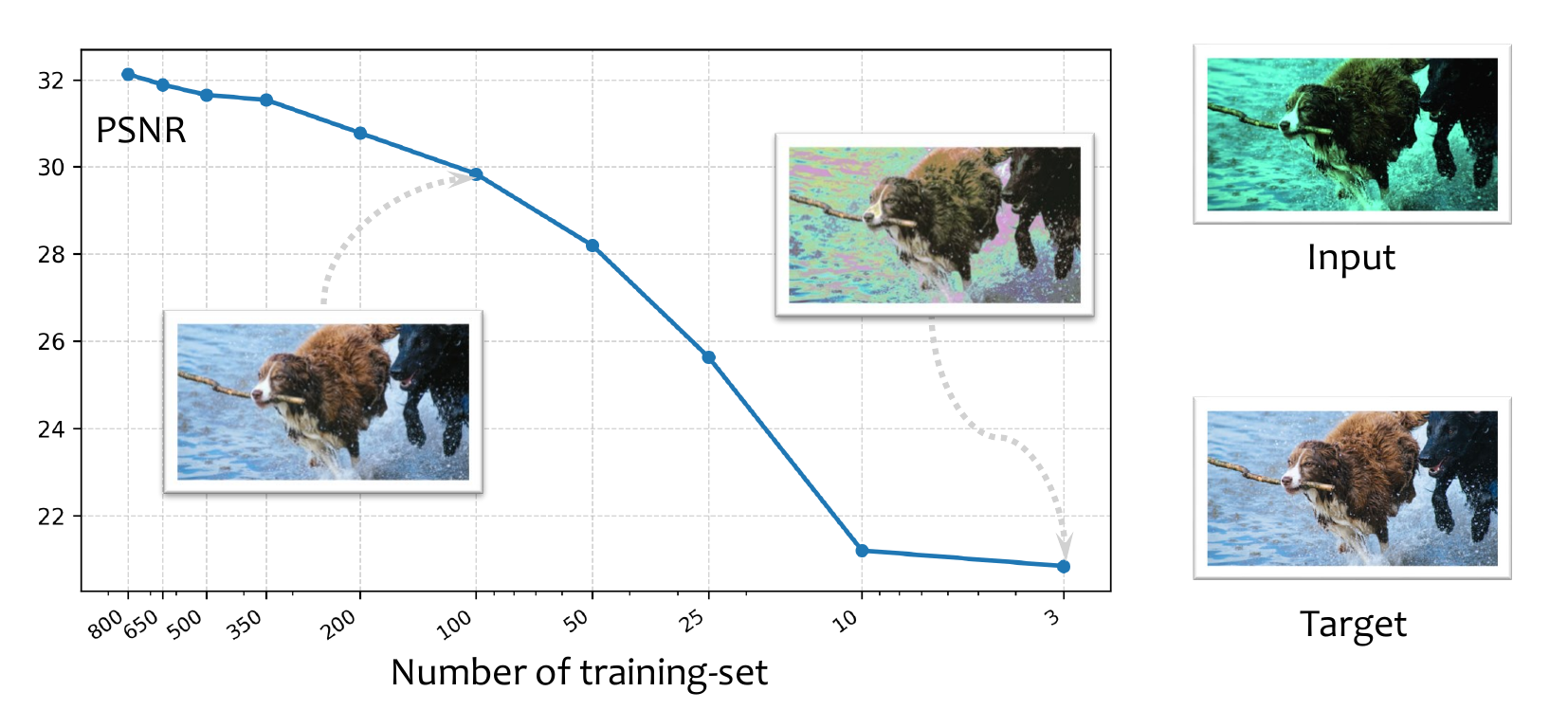}
    \captionof{figure}{Learning efficiency of $\widehat{\mathcal{F}}_{\mathbf{C}}$.}
    \label{fig_12}
\end{minipage}
\hfill
\begin{minipage}{0.54\linewidth}
    \centering
    \includegraphics[width=\linewidth]{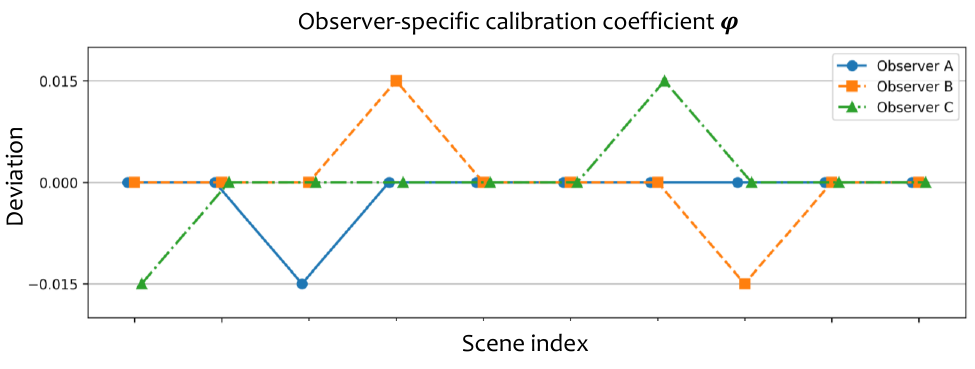}
    \captionof{figure}{Robustness of calibrated $\varphi$.}
    \label{fig_13}
\end{minipage}
\end{figure}

\subsection{Ablations.}
\label{sec_5.3}
\subsubsection{Validity of the DSLR observer proxy.}
We use a DSLR camera as a digital proxy observer for quantitative evaluation. This does not assume that the DSLR is identical to human vision; rather, it serves as a controlled and repeatable three-channel observer, consistent with human's three-cone color responses. As shown in~\cref{fig_11}, DSLR-based preferences follow trends consistent with human-subject judgments across diverse scenes, while often providing a stricter evaluation.

\subsubsection{Learning Efficiency of $\widehat{\mathcal{F}}_{\mathbf{C}}$.} 
We test the learned $\widehat{\mathcal{F}}_{\mathbf{C}}$ with limited training data. We progressively subsample the training set and find that even with 100 images for training, the results still preserves similar colors, as shown in~\cref{fig_12}.

\subsubsection{Robustness of Calibrated $\varphi$.}
To assess whether the calibrated vector $\varphi$ remains stable across illuminants for the same observer $\mathbf{M}$, we present a user study by perturbing each entry of $\varphi$ by $\pm 0.015$, and ask the participant whether any of these alternatives better match the real scene. We visualize the selections of three representative observers across 10 scenes in~\cref{fig_13}. For most scenes, observers keep their original $\varphi$ without switching, indicating cross-illuminant stability.

% To assess whether the calibrated vector $\varphi$ remains stable across illuminants for the same observer $\mathbf{M}$, we include an additional robustness check during the user study, we present a small set of alternative coefficients by perturbing each entry of $\varphi$ by $\pm 0.015$, and ask the participant whether any of these alternatives better match the real scene under the current illumination. If so, the participant re-selects the preferred coefficient.

% We visualize the selections of three representative observers across 10 scenes in Figure~\ref{fig_5.3}. For most scenes, observers keep their original $\varphi$ without switching, indicating cross-illuminant stability. When a change is selected, it exhibits a consistent trend: a slight increase in the $R$ entry and a slight decrease in the $B$ entry. Together, these results support that $\varphi$ is relatively robust across different illuminants for a fixed observer $\mathbf{M}$.

% \subsection{Applications}
% \label{sec_5.4}

% \paragraph{Combined with Camera ISP.}

% \paragraph{Multi-illuminant}

% \paragraph{Similarity Between Camera and Human}

\section{Conclusion}
We presented \textbf{Color Pass-Through}, an end-to-end learned correction applied to captured images via a coupled camera–display pair. 
After optimizing two predictors, we combine them at inference time and perform a one-time coefficient calibration for a target observer. 
This enables color pass-through across diverse scenes and illuminants, as confirmed by both evaluation metrics and user studies.

% \section{Limitations}
% Our method requires optimizing a projector for each specific camera--display pair, rather than learning a single generative model that generalizes across arbitrary devices; due to display’s limited dynamic range, under extreme illumination (e.g., direct sunlight), faithful color reproduction may require HDR capture and rendering to avoid clipping and compression of bright regions.

\clearpage  % TODO FINAL: This \clearpage needs to be removed from both review and camera-ready versions.

\section*{Acknowledgements}
This work was supported in part by the Research Grants Council of Hong Kong under the Early Career Scheme (ECS), Grant No. 24209224.

\bibliographystyle{splncs04}
\bibliography{main}

% ----------------------------------------------------------------
% Supplementary material.  This is included in the same source file so
% arXiv builds one PDF containing both the paper and the supplement.
\clearpage
\appendix
\section*{Supplementary Material}
\addcontentsline{toc}{section}{Supplementary Material}
In this supplementary document, we provide five sections that extend the main paper on our proposed \emph{Color Pass-Through} method.
\begin{itemize}
    \item \cref{supple-sec-1}: We describe the \textbf{model in detail}, including the full implementation and key component designs.
    \item \cref{Detailed_Mathematical_Proof}: We present \textbf{detailed mathematical derivations}. We first derive the full model in the linear case, and then extend the camera--display projector to the nonlinear setting together with the low-rank approximation of camera-null correction, leading to the final formulation.
    \item \cref{supple-sec-3}: We provide \textbf{extra experiments}, including evaluating the effectiveness of AvgPool augment and comparing various baselines for learning \(\mathcal{F}_{\mathbf C}\).
    \item \cref{supple-sec-4}: We discuss \textbf{extended applications} of Color Pass-Through, including its compatibility with in-phone camera processing for producing results with richer details and lower noise, as well as its extension to high-dynamic-range scenes for alleviating overexposure in highlight regions.
    \item \cref{supple-sec-5}: We discuss the \textbf{limitations} of the current version and outline potential directions for future improvement.
\end{itemize}

\section{Model Details}
\label{supple-sec-1}
\subsection{Full Model Implementation}
\begin{figure}[t]
\centering
  \includegraphics[width=\linewidth]{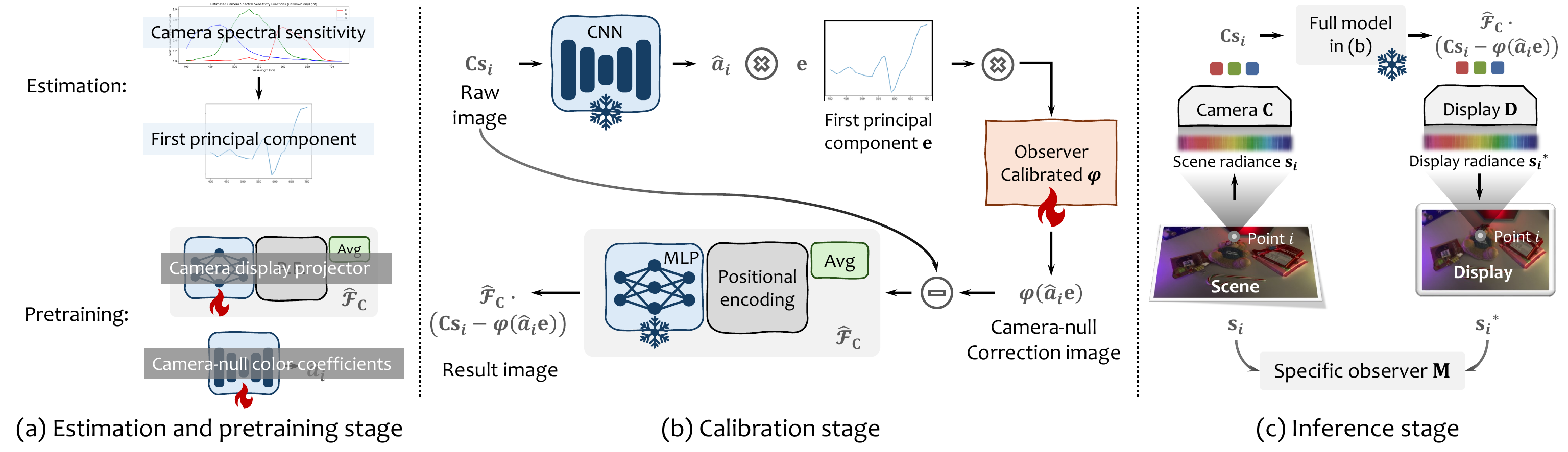}
  \caption{\textbf{Three Stages of Color Pass-Through.} (a) In the estimation and pretraining stage, we first estimate the camera spectral sensitivity \(\mathbf{C}\) and use it to derive the first principal component \(\mathbf{e}\), after which the two predictors are trained separately. (b) In the calibration stage, the full model is executed once using the two pretrained predictors to estimate the calibration coefficient $\varphi$. (c) In the inference stage, we evaluate whether the scene color $\mathbf{M}\mathbf{s}_i$ is consistent with the display color $\mathbf{M}\mathbf{s}^{*}_i$ for a given observer $\mathbf{M}$.}
  \label{supple_fig_1}
\end{figure}

Our proposed Color Pass-Through framework consists of three stages: a pretraining stage, a calibration stage, and an inference stage, as illustrated in~\cref{supple_fig_1}.

\begin{itemize}
    \item \textbf{Estimation and pretraining stage.} This stage consists of two steps. 
    (i) We first estimate the camera sensitivity function $\mathbf{C}$ from ColorChecker images using~\cite{jiang2013space}, and then use the estimated $\mathbf{C}$ to derive the first principal component $\mathbf{e}$ of the camera null-space basis from hyperspectral data. 
    (ii) We independently pretrain two predictor networks. The first is a learned camera--display projector, $\widehat{\mathcal{F}}_{\mathbf{C}}(\cdot): \mathbb{R}^{3}\rightarrow\mathbb{R}^{3}$, for a coupled camera $\mathbf{C}$ and display $\mathbf{D}$, which enables general pass-through. It yields accurate results in the special case of $\mathbf{M}=\mathbf{C}$, while a residual color discrepancy remains when $\mathbf{M}\neq\mathbf{C}$. The second predictor learns the camera-null color coefficients $\widehat{a}_i \in \mathbb{R}$, which estimate the metameric-black component from raw images and provide the basis for camera-null color correction for a specific observer $\mathbf{M}$.

    \item \textbf{Calibration stage.} We combine the two pretrained predictors and capture a raw image of an arbitrary scene using the phone's Pro mode. We then perform a 3D-grid-based calibration procedure to estimate the calibration coefficient $\varphi \in \mathbb{R}^{3 \times 1}$. This coefficient is computed once and then fixed for the target observer, whether a human observer or a digital observer.

    \item \textbf{Inference stage.} Finally, we combine the two pretrained predictors with the calibrated coefficient $\varphi$ to perform validation and evaluation.
\end{itemize}

\newpage

\subsection{Estimation and Pretraining}

\subsubsection{Estimating camera spectral sensitivity $\mathbf{C}$.}
\label{sec_1.2.1}
We first estimate $\mathbf{C}$ in order to derive the first principal component $\mathbf{e}$ described in next section. 
Following Jiang et al.~\cite{jiang2013space}, we model the camera spectral sensitivity functions over the visible range (400--720\,nm) using a low-dimensional PCA basis. Let $\mathbf{C}=[C_R;C_G;C_B]\in\mathbb{R}^{3\times L}$, where $C_k$ denotes the spectral sensitivity of channel $k\in\{R,G,B\}$ sampled at $L$ wavelengths. For a ColorChecker patch $j$ with known spectral reflectance $R_j(\lambda)$ (we adopt from~\cite{beep6581_dcamprof}) under an illuminant spectrum $L(\lambda)$, the recorded response is given by:
\[
I_{j,k}=\int C_k(\lambda)\,L(\lambda)\,R_j(\lambda)\,d\lambda,
\]
or, in discretized form,
\[
I_{j,k}\approx \Delta\lambda\sum_{\lambda} C_k(\lambda)\,L(\lambda)\,R_j(\lambda).
\]

As in~\cite{jiang2013space}, we normalize the spectral sensitivity of each channel and represent it in a PCA subspace learned from measured camera sensitivities. Specifically, the normalized sensitivity of channel $k$ is written as:
\[
\mathbf{c}_{k,n}=\boldsymbol{\sigma}_k \mathbf{E}_k,
\]
where $\boldsymbol{\sigma}_k\in\mathbb{R}^{1\times 2}$ denotes the PCA coefficients and $\mathbf{E}_k\in\mathbb{R}^{2\times L}$ contains the first two principal components for that channel. The full sensitivity is then given by:
\[
\mathbf{c}_k=g_k\,\mathbf{c}_{k,n}=g_k\,\boldsymbol{\sigma}_k\mathbf{E}_k,
\]
where $g_k$ is a channel-dependent gain. Jiang et al.~\cite{jiang2013space} show that the first two principal components explain more than 97\% of the variance; therefore, a two-dimensional model is sufficient for each RGB channel.

When the illuminant is unknown but assumed to be daylight, the illuminant spectrum is modeled using the CIE daylight basis~\cite{judd1964spectral}:
\[
L(\lambda,t)=\bar{L}(\lambda)+M_1(t)V_1(\lambda)+M_2(t)V_2(\lambda),
\]
where $\bar{L}(\lambda)$ is the mean daylight spectrum, $V_1(\lambda)$ and $V_2(\lambda)$ are the characteristic daylight basis vectors, and $M_1(t)$ and $M_2(t)$ are determined by the correlated color temperature (CCT) $t$. Under this model, both the daylight parameter $t$ and the PCA coefficients $\boldsymbol{\sigma}_k$ are optimized iteratively to minimize the forward-model error between the measured average Color-checker patch images and the responses predicted by the recovered sensitivities.

\begin{figure}[t]
\centering
  \includegraphics[width=\linewidth]{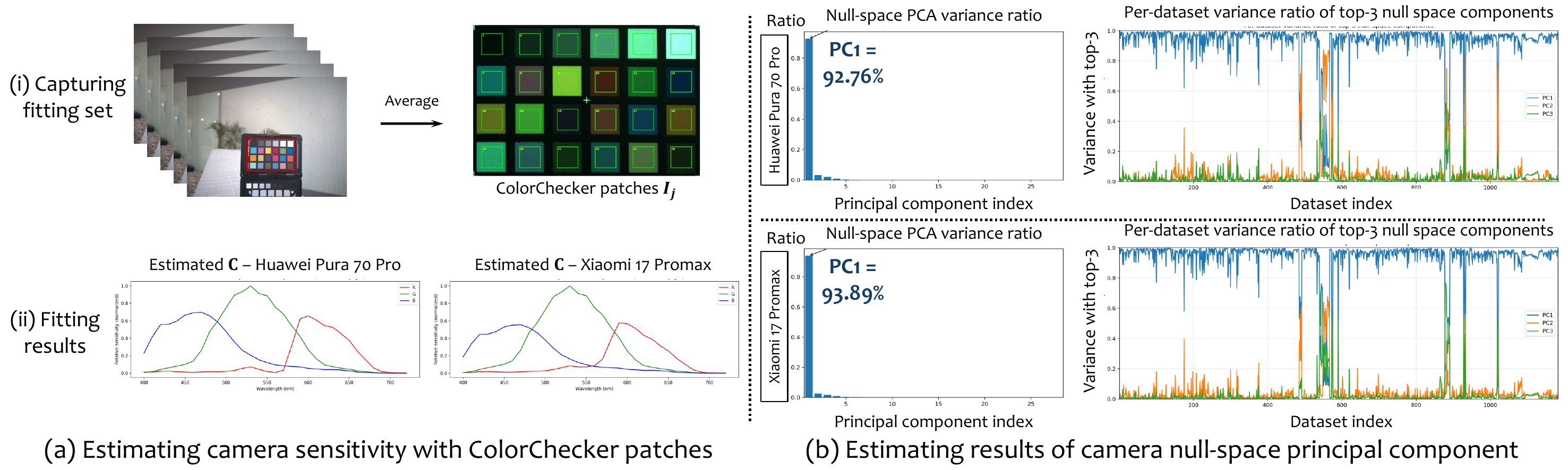}
    \caption{\textbf{Estimating Results of Camera Sensitivity and Null-Space Components.} 
    (a) Capture setup and estimated camera sensitivities. 
    (b) PCA variance ratios of the estimated camera null space over the full dataset (left), and variance ratios of the top-3 components for each sample (right), PC1 denotes the first principal component.}
  \label{supple_fig_2}
\end{figure}

In our implementation, we capture 15 ColorChecker images under daylight illumination and average the estimated results, as shown in~\cref{supple_fig_2}~(a).

\subsubsection{Estimating the first principal component $\mathbf{e}$.}
\label{sec_1.2.2}

Let $\mathbf{C}\in\mathbb{R}^{L\times 3}$ denote the estimated camera sensitivity in previous section, aligned with the hyperspectral sampling grid, where $L=31$ spectral bands. We then estimate the camera null-space components, with particular focus on the first principal component $\mathbf{e}$. Specifically, we first compute an orthonormal basis $\mathbf{N}\in\mathbb{R}^{L\times d}$ for $\mathrm{null}(\mathbf{C}^{\top})$, where $d=L-\mathrm{rank}(\mathbf{C})=28$. For a spectrum $\mathbf{s}\in\mathbb{R}^{L}$, its camera-invisible (metameric-black) component~\cite{morovivc2006metamer,alsam2006calibrating,vienot2013verriest} is given by the orthogonal projection:
\[
\mathbf{s}_{\mathrm{null}}=\mathbf{N}\mathbf{N}^{\top}\mathbf{s}.
\]
We then represent $\mathbf{s}_{\mathrm{null}}$ in null-space coordinates as:
\[
\mathbf{z}=\mathbf{s}^{\top}\mathbf{N}\in\mathbb{R}^{d},
\]
and perform PCA in this reduced space. Concretely, we estimate the covariance of $\mathbf{z}$ over the selected hyperspectral fitting set, including ARAD-1K~\cite{arad2022ntire}, CAVE~\cite{yasuma2010generalized}, and ICVL~\cite{arad2016sparse}, and compute its eigendecomposition. The leading eigenvectors then define the principal directions of variation in the null space.

Across 1,182 hyperspectral images ($6.02\times 10^{8}$ pixels in total), as shown in~\cref{supple_fig_2}~(b), we observe that the null-space distribution is highly concentrated along the first principal direction. The first principal component explains over $90\%$ of the total null-space variance, while the first three components explain over $98\%$ cumulatively. In addition, reconstruction using the top three components preserves nearly $99\%$ of the total energy. These results indicate that, although the null space is theoretically 28-dimensional, its effective dimensionality is close to one for our data. Therefore, in the following, we use only the first principal basis vector, denoted by $\mathbf{e}$, to parameterize the null-space component.

% \subsubsection{Pretraining Individual Models}
% We show the detail training process in the main text of our paper in sec 4.1 and sec 4.2 respectively,  
% trained two models using one NVIDIA 4090

\begin{figure}[t]
\centering
  \includegraphics[width=\linewidth]{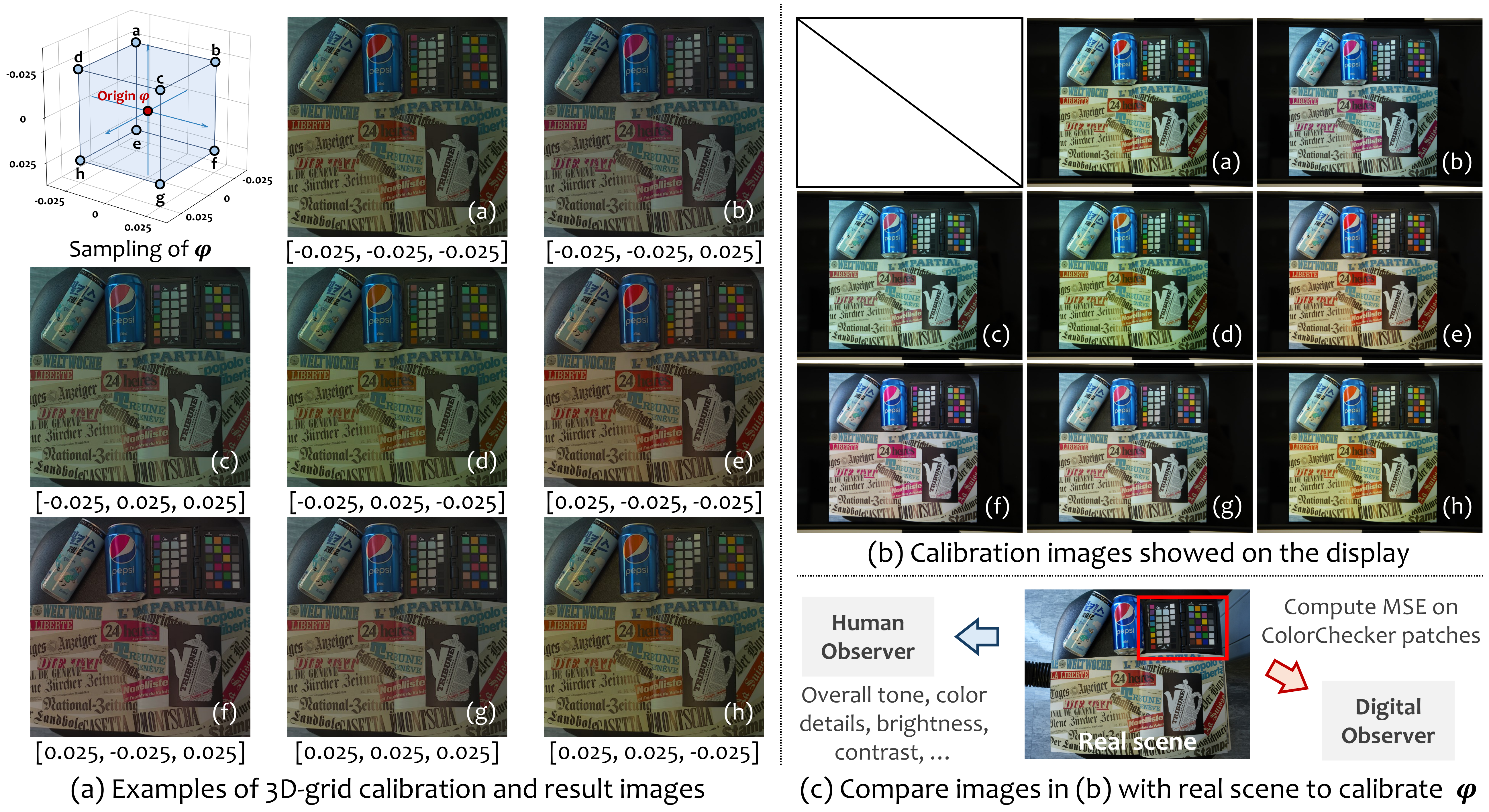}
  \setlength{\abovecaptionskip}{-0.1cm}
  \setlength{\belowcaptionskip}{-0.3cm}
    \caption{\textbf{Observer-Specific 3D Palette Calibration.} 
    (a) A one-time calibration is performed in an arbitrary scene by sampling \(\varphi\) in a 3D-grid space. 
    (b) The sampled results are showed on the display. 
    (c) The observer selects the one with \(\varphi\) that best matches the real scene, which is then fixed for all subsequent experiments.}
  \label{supple_fig_3}
\end{figure}

\subsection{Observer-Specific Calibration via a 3D-grid}
After estimation and pretraining, we combine the two predictors and run the full Color Pass-Through model once to calibrate the coefficient \(\varphi\in\mathbb{R}^{3\times 1}\), as illustrated in~\cref{supple_fig_1}(b). As shown in the derivation of~\cref{eq:supp_delta_pca_expand}, under the low-rank approximation, \(\varphi\) is independent of the scene radiance \(\mathbf{s}_i\). Moreover, Fig.~12 in the main paper shows that \(\varphi\) remains nearly stable across diverse real-world scenes. Therefore, we calibrate \(\varphi\) only once for each observer using an arbitrary scene and keep it fixed throughout all subsequent experiments.

% We adopt a 3D-grid calibration for \(\varphi\). We first generate the result image under various \(\varphi\) settings (8 examples shows in~\cref{supple_fig_3}~(a)), then show them on the display (\cref{supple_fig_3}~(b)), lastly we let the observer to choose the best \(\varphi\). For human observer, they will compare from the overall tone to detailed color; for digital observer (DSLR), we perform MSE loss on the ColorChecker patches.

We adopt a 3D-grid calibration for \(\varphi\). Specifically, we first generate result images under different \(\varphi\) (eight examples shown in~\cref{supple_fig_3}(a)), and then render them on the display (\cref{supple_fig_3}(b)). The observer then selects the value of \(\varphi\) that provides the best match (\cref{supple_fig_3}(c)). For a human observer, the selection is based on both the overall tone and fine-grained color details. For a digital observer (DSLR), we determine \(\varphi\) by minimizing MSE over the ColorChecker patches.

\section{Detailed Mathematical Proof}
\label{Detailed_Mathematical_Proof}
\begin{figure}[t]
\centering
  \includegraphics[width=\linewidth]{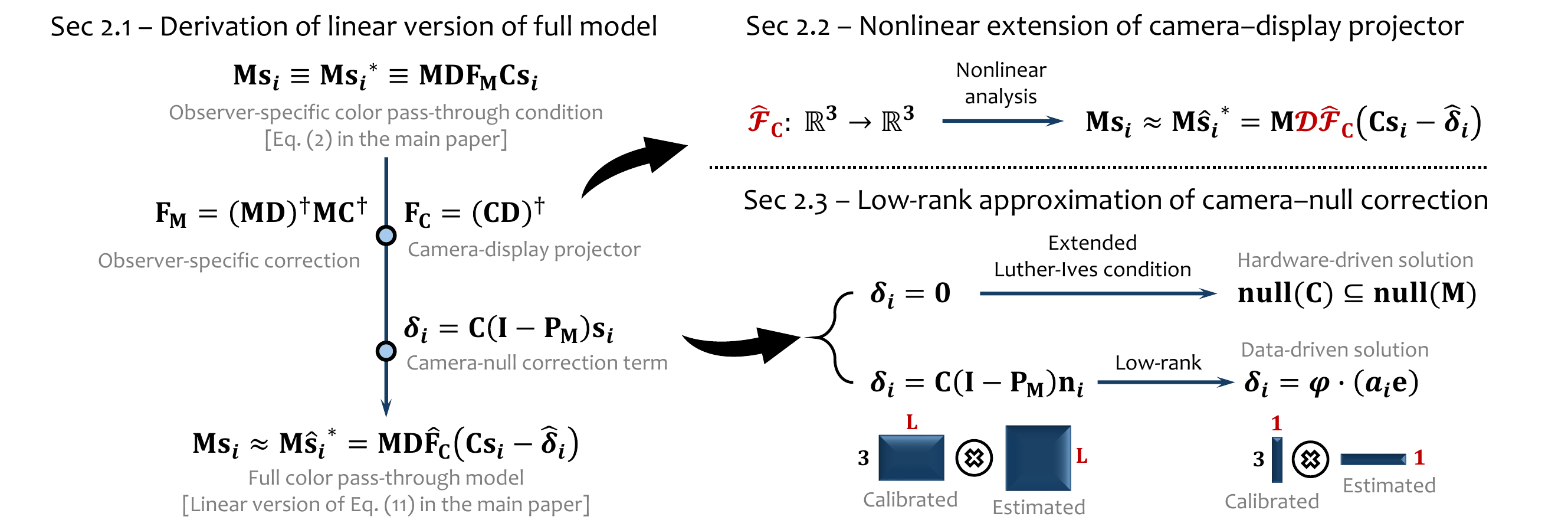}
    \caption{\textbf{Schematic Overview of Detailed Mathematical Proof in~\cref{Detailed_Mathematical_Proof}.}}
  \label{supple_fig_4}
\end{figure}

We provide a schematic flowchart of the derivation, together with the key conditions, in~\cref{supple_fig_4}. Readers are encouraged to refer to it throughout this section.

\subsection{Review of Full Model Interpretation in the Linear Case}
\label{supp_review_full_model_linear}

In this section, we first derive Eq.~(5) and then present the linear formulation of the full model in Eq.~(11) of the main paper. Our goal is to show that the observer-aware mapping \(\mathbf{F}_{\mathbf M}\) can be interpreted as the camera--display projector \(\mathbf{F}_{\mathbf C}\) augmented with a correction term \(\delta_i \in \mathbb{R}^{3}\) in the input space of \(\mathbf{F}_{\mathbf C}\).

Starting from Eq.~(2) in the main text, the observer-preserving condition is:
\begin{equation}
\mathbf{M}\mathbf{s}_i
\;\equiv\;
\mathbf{M}\mathbf{s}_i^{*}
\;\equiv\;
\mathbf{M}\mathbf{D}\mathbf{F}_{\mathbf M}\mathbf{C}\mathbf{s}_i,
\quad \forall i,
\label{eq:supp_1}
\end{equation}
which gives:
\begin{equation}
\mathbf{F}_{\mathbf M}
=
(\mathbf{M}\mathbf{D})^{\dagger}\mathbf{M}\mathbf{C}^{\dagger},
\label{eq:supp_fm}
\end{equation}
Multiplying both sides by \(\mathbf{C}\mathbf{s}_i\), and using the standard pseudo-inverse property
\(
\mathbf{C}^{\dagger}\mathbf{C}\mathbf{s}_i=\mathbf{s}_i
\)
on the identifiable signal subspace, we obtain:
\begin{equation}
\mathbf{F}_{\mathbf M}\mathbf{C}\mathbf{s}_i
=
(\mathbf{M}\mathbf{D})^{\dagger}\mathbf{M}\mathbf{s}_i,
\label{eq:supp_fmcs}
\end{equation}

Next, define the observer-induced spectral projector:
\begin{equation}
\mathbf{P}_{\mathbf M}
:=
\mathbf{D}(\mathbf{M}\mathbf{D})^{\dagger}\mathbf{M},
\label{eq:supp_pm}
\end{equation}
By construction,
\begin{equation}
\mathbf{C}\mathbf{P}_{\mathbf M}\mathbf{s}_i
=
\mathbf{C}\mathbf{D}(\mathbf{M}\mathbf{D})^{\dagger}\mathbf{M}\mathbf{s}_i,
\label{eq:supp_cpms}
\end{equation}
Applying the camera--display projector:
\begin{equation}
\mathbf{F}_{\mathbf C}
=
(\mathbf{C}\mathbf{D})^{\dagger},
\label{eq:supp_fc}
\end{equation}
to both sides yields:
\begin{equation}
\mathbf{F}_{\mathbf C}\mathbf{C}\mathbf{P}_{\mathbf M}\mathbf{s}_i
=
(\mathbf{C}\mathbf{D})^{\dagger}\mathbf{C}\mathbf{D}(\mathbf{M}\mathbf{D})^{\dagger}\mathbf{M}\mathbf{s}_i,
\label{eq:supp_fc_cpms}
\end{equation}
Again using the pseudo-inverse identity
\(
(\mathbf{C}\mathbf{D})^{\dagger}\mathbf{C}\mathbf{D}=\mathbf{I}
\)
on the corresponding display-code subspace, we have:
\begin{equation}
\mathbf{F}_{\mathbf C}\mathbf{C}\mathbf{P}_{\mathbf M}\mathbf{s}_i
=
(\mathbf{M}\mathbf{D})^{\dagger}\mathbf{M}\mathbf{s}_i,
\label{eq:supp_fc_pm_result}
\end{equation}
Combining \cref{eq:supp_fmcs,eq:supp_fc_pm_result} gives:
\begin{equation}
\mathbf{F}_{\mathbf M}\mathbf{C}\mathbf{s}_i
=
\mathbf{F}_{\mathbf C}\mathbf{C}\mathbf{P}_{\mathbf M}\mathbf{s}_i,
\label{eq:supp_bridge}
\end{equation}

Finally, define the residual term:
\begin{equation}
\delta_i
:=
\mathbf{C}\bigl(\mathbf{I}-\mathbf{P}_{\mathbf M}\bigr)\mathbf{s}_i,
\label{eq:supp_delta}
\end{equation}
Then:
\begin{equation}
\mathbf{C}\mathbf{P}_{\mathbf M}\mathbf{s}_i
=
\mathbf{C}\mathbf{s}_i-\delta_i,
\label{eq:supp_cpms_delta}
\end{equation}
Substituting \cref{eq:supp_cpms_delta} into \cref{eq:supp_bridge}, we obtain:
\begin{equation}
\mathbf{F}_{\mathbf M}\mathbf{C}\mathbf{s}_i
\;=\;
\mathbf{F}_{\mathbf C}\bigl(\mathbf{C}\mathbf{s}_i-\delta_i\bigr),
\qquad
\delta_i
\;:=\;
\mathbf{C}\bigl(\mathbf{I}-\mathbf{P}_{\mathbf M}\bigr)\mathbf{s}_i,
\label{eq:supp_full_interpretation}
\end{equation}
\textbf{which is exactly the Eq.~(5) in the main paper.}

Combining~\cref{eq:supp_full_interpretation} with the full-model interpretation~\cref{eq:supp_1} yields the full Color Pass-through model (linear version) used in the main paper:
\begin{equation}
\underbrace{\mathbf{M}\mathbf{s}_i}_{\text{scene color}}
\approx
\underbrace{\mathbf{M}\widehat{\mathbf{s}}_i^{*}}_{\text{display color}}
=
\mathbf{M}\mathbf{D}\cdot
\underbrace{\widehat{\mathbf{F}}_{\mathbf C}}_{\textbf{learned}}
\cdot
\Bigl(
\underbrace{\mathbf{C}\mathbf{s}_i}_{\text{camera color}}
-
\underbrace{\widehat{\delta_i}}_{\textbf{learned}}
\Bigr),
\label{eq:supp_full_model_final}
\end{equation}

\textbf{This is exactly the linear version of Eq.~(11) in the main paper.} This derivation provides an intuitive interpretation of the full linear model: we first correct the camera color \(\mathbf{C}\mathbf{s}_i\) by subtracting the residual color term \(\widehat{\delta_i}\), and then apply the camera--display projector \(\widehat{\mathbf{F}}_{\mathbf C}\). The resulting signal is rendered on display \(\mathbf{D}\) and, when observed by \(\mathbf{M}\), yields a display color \(\mathbf{M}\widehat{\mathbf{s}}_i^{*}\) that closely matches the scene color \(\mathbf{M}\mathbf{s}_i\).

\subsection{Nonlinear Extension of the Camera--Display Projector}
\label{supp_nonlinear_fc}

The derivation above is presented under a linear formulation, where the camera--display projector is written as
\(
\mathbf{F}_{\mathbf C}=(\mathbf{C}\mathbf{D})^{\dagger}
\).
In practice, however, the effective display $\mathbf{D}$ is not strictly linear due to gamma correction, tone mapping, and other device-dependent nonlinear processing. We therefore generalize the camera--display projector $\mathbf{F}_{\mathbf C}$ to a nonlinear operator $\mathcal{F}_{\mathbf C}$ and show that the full model in Eq.~\eqref{eq:supp_full_model_final} remains valid in the same compositional form.

\paragraph{(1) Nonlinear camera--display forward model.}
Instead of modeling the display by a linear matrix \(\mathbf{D}\), we introduce a nonlinear display emission operator:
\begin{equation}
\mathcal{D}:\mathbb{R}^{3}\rightarrow\mathbb{R}^{L},
\end{equation}
which maps a digital RGB signal \(\sigma\in\mathbb{R}^{3}\) to its emitted spectrum. The corresponding effective camera--display forward process is then represent as:
\begin{equation}
\mathcal{T}_{\mathbf{C}\mathbf{D}}(\sigma)
:=
\mathbf{C}\,\mathcal{D}(\sigma),
\qquad
\mathcal{T}_{\mathbf{C}\mathbf{D}}:\mathbb{R}^{3}\rightarrow\mathbb{R}^{3}.
\label{eq:supp_nonlinear_forward}
\end{equation}
This operator replaces the linear mapping \(\mathbf{C}\mathbf{D}\).

\paragraph{(2) Nonlinear camera--display projector.}
We now define the nonlinear camera--display projector as a learned operator:
\begin{equation}
\mathcal{F}_{\mathbf C}:\mathbb{R}^{3}\rightarrow\mathbb{R}^{3},
\label{eq:supp_nonlinear_fc_def}
\end{equation}
which is required to act as a right inverse of the effective forward process on the target camera-color domain, i.e.,
\begin{equation}
\mathcal{T}_{\mathbf{C}\mathbf{D}}\bigl(\mathcal{F}_{\mathbf C}(\mathbf{c})\bigr)
=
\mathbf{c},
\qquad
\forall \mathbf{c}\in\Omega_{\mathbf C},
\label{eq:supp_nonlinear_right_inverse}
\end{equation}
where \(\Omega_{\mathbf C}\subseteq\mathbb{R}^{3}\) denotes the set of realizable camera colors. Equivalently,
\begin{equation}
\mathbf{C}\,\mathcal{D}\bigl(\mathcal{F}_{\mathbf C}(\mathbf{c})\bigr)
=
\mathbf{c},
\label{eq:supp_nonlinear_right_inverse_expanded}
\end{equation}
Thus, \(\mathcal{F}_{\mathbf C}\) generalizes the linear pseudo-inverse \((\mathbf{C}\mathbf{D})^{\dagger}\) to the nonlinear case.

\paragraph{(3) Nonlinear full-model interpretation.}
In the linear derivation above, the observer-aware mapping is interpreted as applying \(\mathbf{F}_{\mathbf C}\) to a corrected camera color \(\mathbf{C}\mathbf{s}_i-\delta_i\). This interpretation does not rely on linearity of the \emph{input correction}; it only requires that the corrected signal lies in the domain on which the camera--display projector is valid. Therefore, once \(\widehat{\delta}_i\) is estimated, we may define the nonlinear displayed radiance as:
\begin{equation}
\widehat{\mathbf{s}}_i^{*}
:=
\mathcal{D}\!\left(
\mathcal{F}_{\mathbf C}\bigl(\mathbf{C}\mathbf{s}_i-\widehat{\delta}_i\bigr)
\right),
\label{eq:supp_nonlinear_sstar}
\end{equation}
Applying the observer \(\mathbf{M}\) gives:
\begin{equation}
\mathbf{M}\widehat{\mathbf{s}}_i^{*}
=
\mathbf{M}\,\mathcal{D}\!\left(
\mathcal{F}_{\mathbf C}\bigl(\mathbf{C}\mathbf{s}_i-\widehat{\delta}_i\bigr)
\right),
\label{eq:supp_nonlinear_observer}
\end{equation}

Accordingly, the full Color Pass-through model takes exactly the same compositional form as in the linear case, except that the camera--display projector is now nonlinear:
\begin{equation}
\underbrace{\mathbf{M}\mathbf{s}_i}_{\text{scene color}}
\approx
\underbrace{\mathbf{M}\widehat{\mathbf{s}}_i^{*}}_{\text{display color}}
=
\mathbf{M}\,\mathcal{D}\!\left(
\underbrace{\widehat{\mathcal{F}}_{\mathbf C}}_{\textbf{learned}}
\Bigl(
\underbrace{\mathbf{C}\mathbf{s}_i}_{\text{camera color}}
-
\underbrace{\widehat{\delta}_i}_{\textbf{learned}}
\Bigr)
\right),
\label{eq:supp_nonlinear_full_model}
\end{equation}

\paragraph{(4) Consistency with the linear form.}
Equation~\eqref{eq:supp_nonlinear_full_model} is the nonlinear counterpart of Eq.~\eqref{eq:supp_full_model_final}. When the display emission is linear, i.e.,
\(
\mathcal{D}(\sigma)=\mathbf{D}\sigma
\),
and \(\widehat{\mathcal{F}}_{\mathbf C}\) is linear, i.e.,
\(
\widehat{\mathcal{F}}_{\mathbf C}(\mathbf{c})=\widehat{\mathbf{F}}_{\mathbf C}\mathbf{c},
\)
Eq.~\eqref{eq:supp_nonlinear_full_model} reduces back to:
\begin{equation}
\mathbf{M}\widehat{\mathbf{s}}_i^{*}
=
\mathbf{M}\mathbf{D}\cdot
\widehat{\mathbf{F}}_{\mathbf C}\cdot
\bigl(\mathbf{C}\mathbf{s}_i-\widehat{\delta}_i\bigr),
\end{equation}
which is exactly Eq.~\eqref{eq:supp_full_model_final}.

\paragraph{(5) Practical form used in the main paper.}
In practice, our learned camera--display projector is implemented as a nonlinear pixel-wise network. Therefore, the model used in the main paper can be interpreted as the practical instantiation of Eq.~\eqref{eq:supp_nonlinear_full_model}: we first correct the camera color \(\mathbf{C}\mathbf{s}_i\) by subtracting the learned residual \(\widehat{\delta}_i\), and then feed the corrected camera color into the learned nonlinear camera--display projector \(\widehat{\mathcal{F}}_{\mathbf C}\). After display emission and observation by \(\mathbf{M}\), the resulting display color \(\mathbf{M}\widehat{\mathbf{s}}_i^{*}\) remains an approximation of the original scene color \(\mathbf{M}\mathbf{s}_i\). \textbf{Therefore, the full-model formulation remains valid even when \(\widehat{\mathbf{F}}_{\mathbf C}\) is generalized from linear pseudo-inverse to nonlinear learned operator.}

\subsection{Low-rank Approximation of Camera--Null Correction}
\label{supp_low_rank_camera_null}

In this section, we provide the derivations omitted in the main text for the correction color term $\delta_i$ in Eq.~(5) of the main text:
\begin{equation}
\delta_i
:=
\mathbf{C}\bigl(\mathbf{I}-\mathbf{P}_{\mathbf M}\bigr)\mathbf{s}_i,
\qquad
\mathbf{P}_{\mathbf M}:=\mathbf{D}(\mathbf{M}\mathbf{D})^{\dagger}\mathbf{M}.
\label{eq:supp_delta_def}
\end{equation}

We first establish a sufficient condition for \(\delta_i=\mathbf{0}\), corresponding to the first regime discussed in the main paper, and provide a hardware-driven route to eliminate the need to estimate \(\delta_i\) when \(\delta_i \approx 0\). We then consider the general case \(\delta_i \neq \mathbf{0}\), and show that \(\delta_i\) arises entirely from the camera-null component of the scene radiance, and thus must be learned. This observation directly motivates the low-rank approximation used in Eq.~(9) of the main text.

\paragraph{(1) A sufficient condition for zero correction $\delta_i=0$.}
From Eq.~(5) in the main paper, zero correction means:
\begin{equation}
\delta_i=\mathbf{0}
\quad\Longleftrightarrow\quad
\mathbf{C}\bigl(\mathbf{I}-\mathbf{P}_{\mathbf M}\bigr)\mathbf{s}_i=\mathbf{0},
\label{eq:supp_zero_delta_start}
\end{equation}
A sufficient condition for this to hold for all \(\mathbf{s}_i\) is:
\begin{equation}
\bigl(\mathbf{I}-\mathbf{P}_{\mathbf M}\bigr)\mathbf{s}_i \in \mathrm{null}(\mathbf C),
\qquad \forall i.
\label{eq:supp_zero_delta_mid}
\end{equation}
We now show that this is guaranteed when:
\begin{equation}
\mathrm{null}(\mathbf C)\subseteq \mathrm{null}(\mathbf M),
\label{eq:supp_luther_ives_extended}
\end{equation}

To see this, recall that \(\mathbf{P}_{\mathbf M}=\mathbf{D}(\mathbf{M}\mathbf{D})^\dagger \mathbf M\) is the oblique projector onto \(\mathrm{range}(\mathbf D)\) along \(\mathrm{null}(\mathbf M)\). Hence, for any spectrum \(\mathbf{s}\):
\begin{equation}
\bigl(\mathbf{I}-\mathbf{P}_{\mathbf M}\bigr)\mathbf{s}\in \mathrm{null}(\mathbf M),
\label{eq:supp_pm_nullm}
\end{equation}
If, in addition, \(\mathrm{null}(\mathbf C)\subseteq \mathrm{null}(\mathbf M)\), then all directions invisible to the camera are also invisible to the observer. Under this condition, whenever the residual component
\(
(\mathbf{I}-\mathbf{P}_{\mathbf M})\mathbf{s}
\)
lies in \(\mathrm{null}(\mathbf C)\), it is automatically annihilated by \(\mathbf M\) as well. Therefore,
\begin{equation}
\mathbf{C}\bigl(\mathbf{I}-\mathbf{P}_{\mathbf M}\bigr)\mathbf{s}=\mathbf 0
\quad\Longrightarrow\quad
\mathbf{M}\bigl(\mathbf{I}-\mathbf{P}_{\mathbf M}\bigr)\mathbf{s}=\mathbf 0,
\label{eq:supp_zero_delta_logic}
\end{equation}
and the observer-aware projection introduces no residual correction in camera space. This proves that \cref{eq:supp_luther_ives_extended} is a sufficient condition for \(\delta_i=\mathbf 0\).

Equivalently, using the camera--display projector \(\mathbf F_{\mathbf C}=(\mathbf C\mathbf D)^\dagger\), we have:
\begin{equation}
(\mathbf C\mathbf D)^\dagger \mathbf C(\mathbf I-\mathbf P_{\mathbf M})\mathbf s = \mathbf 0
\;\Longleftarrow\;
\mathbf C(\mathbf I-\mathbf P_{\mathbf M})\mathbf s = \mathbf 0
\;\Longleftarrow\;
(\mathbf I-\mathbf P_{\mathbf M})\mathbf s \in \mathrm{null}(\mathbf C),
\label{eq:supp_zero_delta_chain}
\end{equation}
Since \((\mathbf I-\mathbf P_{\mathbf M})\mathbf s\in \mathrm{null}(\mathbf M)\) by construction, this residual lies in:
\begin{equation}
(\mathbf I-\mathbf P_{\mathbf M})\mathbf s \in \mathrm{null}(\mathbf C)\cap \mathrm{null}(\mathbf M),
\label{eq:supp_intersection}
\end{equation}
which characterizes the ideal case in which the correction term $\delta_i$ vanishes. 

\paragraph{(2) Extended Luther--Ives condition.}
Based on~\cref{eq:supp_intersection}, we obtain a sufficient condition for \(\delta_i=\mathbf{0}\) to hold for all scene radiances \(\mathbf{s}_i\), namely~\cref{eq:supp_luther_ives_extended}:
\[
\mathrm{null}(\mathbf{C}) \subseteq \mathrm{null}(\mathbf{M}),
\]
This follows because \((\mathbf{I}-\mathbf{P}_{\mathbf{M}})\mathbf{s}\in \mathrm{null}(\mathbf{M})\) by construction. Therefore, if every direction in the camera null space is also contained in the observer null space, then any residual component invisible to the camera is necessarily invisible to the observer as well. In the special case where:
\begin{equation}
\mathbf M = \mathbf T \mathbf C,
\qquad \mathbf T \in \mathbb R^{3\times 3}\ \text{invertible},
\label{eq:supp_mtc}
\end{equation}
we have
\begin{equation}
\mathrm{null}(\mathbf C)=\mathrm{null}(\mathbf M),
\label{eq:supp_null_equal}
\end{equation}
which corresponds exactly to the classical Luther--Ives condition up to an invertible linear transform. Under this condition, the observer and the camera share the same null space, and the correction term therefore vanishes identically. We refer to this as the \emph{extended Luther--Ives condition under camera--display projection}, as it guarantees zero correction in our projector-based formulation.

This perspective also provides an intuitive explanation for practical differences across devices. In principle, cameras are designed to approximate a colorimetric observer. In practice, however, their spectral sensitivities may deviate substantially from this ideal due to manufacturing limitations. DSLR cameras may in practice be closer to satisfying \cref{eq:supp_luther_ives_extended} than smartphone cameras, which may partly explain why smartphone imaging pipelines often struggle to preserve the perceived ``cinematic'' color of real scenes despite aggressive post-processing.

More importantly, \cref{eq:supp_luther_ives_extended} suggests a hardware-driven route toward enforcing \(\delta_i=0\). Increasing the number of camera channels reduces \(\mathrm{null}(\mathbf C)\), thereby making the inclusion
\[
\mathrm{null}(\mathbf C)\subseteq \mathrm{null}(\mathbf M)
\]
easier to satisfy. This suggests that multi-channel capture systems, such as RGBW or other non-Bayer four-channel CFA designs (e.g., yellow-inclusive patterns), as well as more general multispectral sensors, could reduce or even eliminate the correction term by design \cite{Prasad2016_3plus1,Anzagira2015_CFA,Oh2017_RGBW,Jee2018_RGBW,Brauers2006_MSFA,Nystrom2006_Multispectral}. In other words, adding sensing channels provides a principled hardware-level strategy for improving color pass-through: as the camera null space becomes smaller, fewer observer-relevant spectral variations are lost during capture, and no need for correcting $\delta_i$ as $\delta_i\approx0$.

\paragraph{(3) Decomposition into displayable and camera-null components.}
We now consider the general case, where \(\delta_i\neq \mathbf 0\) for a specific observer \(\mathbf M\). Following Eq.~(7) in the main paper, we decompose the scene radiance as:
\begin{equation}
\mathbf{s}_i=\mathbf r_i+\mathbf n_i,
\qquad
\left\{
\begin{aligned}
\mathbf r_i &:= \mathbf P_{\mathbf C}\mathbf s_i \in \mathrm{range}(\mathbf D),\\
\mathbf n_i &:= \mathbf s_i-\mathbf r_i \in \mathrm{null}(\mathbf C),
\end{aligned}
\right.
\label{eq:supp_decompose}
\end{equation}
where
\begin{equation}
\mathbf P_{\mathbf C}:=\mathbf D(\mathbf C\mathbf D)^\dagger \mathbf C,
\label{eq:supp_pc}
\end{equation}
By construction, \(\mathbf r_i\) is the displayable component of \(\mathbf s_i\), while \(\mathbf n_i\) is the metameric-black component invisible to the camera.

Substituting \(\mathbf s_i=\mathbf r_i+\mathbf n_i\) into \cref{eq:supp_delta_def} gives:
\begin{equation}
\delta_i
=
\mathbf C(\mathbf I-\mathbf P_{\mathbf M})\mathbf r_i
+
\mathbf C(\mathbf I-\mathbf P_{\mathbf M})\mathbf n_i,
\label{eq:supp_delta_split}
\end{equation}
We now show that the first term vanishes. Since \(\mathbf r_i\in \mathrm{range}(\mathbf D)\), there exists some \(\sigma_i\in\mathbb R^3\) such that:
\begin{equation}
\mathbf r_i=\mathbf D\sigma_i,
\label{eq:supp_r_in_D}
\end{equation}
Then
\begin{align}
\mathbf C(\mathbf I-\mathbf P_{\mathbf M})\mathbf r_i
&=
\mathbf C\mathbf r_i-\mathbf C\mathbf P_{\mathbf M}\mathbf r_i \notag\\
&=
\mathbf C\mathbf D\sigma_i-\mathbf C\mathbf D(\mathbf M\mathbf D)^\dagger \mathbf M\mathbf D\sigma_i,
\label{eq:supp_r_term}
\end{align}
Because \((\mathbf M\mathbf D)^\dagger(\mathbf M\mathbf D)\) acts as identity on the display-code subspace, we have:
\begin{equation}
\mathbf P_{\mathbf M}\mathbf r_i=\mathbf r_i,
\label{eq:supp_pm_fix_r}
\end{equation}
and therefore:
\begin{equation}
\mathbf C(\mathbf I-\mathbf P_{\mathbf M})\mathbf r_i=\mathbf 0,
\label{eq:supp_r_vanish}
\end{equation}
Hence, \cref{eq:supp_delta_split} simplifies to
\begin{equation}
\delta_i
=
\mathbf C(\mathbf I-\mathbf P_{\mathbf M})\mathbf n_i,
\label{eq:supp_delta_from_n}
\end{equation}
\textbf{This proves the statement in the main text:} the correction term originates entirely from the camera-null component \(\mathbf n_i\), rather than from the displayable component \(\mathbf r_i\).

\paragraph{(4) Why a low-rank approximation is natural.}
Equation~\eqref{eq:supp_delta_from_n} indicates that estimating \(\delta_i\) requires access to the unobserved spectral component \(\mathbf n_i\in\mathbb R^L\), together with the high-dimensional operator
\[
\mathbf C(\mathbf I-\mathbf P_{\mathbf M})\in\mathbb R^{3\times L}.
\]
Directly recovering \(\mathbf n_i\) or calibrating the full operator is impractical. We therefore leverage the classical observation that natural spectra are typically well approximated by a low-dimensional subspace~\cite{maloney1986evaluation,parkkinen1989characteristic}. Accordingly, we assume that (also supported by the previous experimental evidence in~\cref{supple_fig_2}) the camera-null component can also be represented in a low-rank basis:
\begin{equation}
\mathbf n_i \approx \sum_{k=1}^{K} a_i^{(k)} \mathbf e^{(k)},
\qquad K\ll L,
\label{eq:supp_n_pca}
\end{equation}
where \(\{\mathbf e^{(k)}\}_{k=1}^{K}\) are basis vectors spanning the dominant modes of variation in camera-null spectra.

Substituting \cref{eq:supp_n_pca} into \cref{eq:supp_delta_from_n} yields
\begin{align}
\delta_i
&=
\mathbf C(\mathbf I-\mathbf P_{\mathbf M})\mathbf n_i
\notag\\
&\approx
\mathbf C(\mathbf I-\mathbf P_{\mathbf M})
\sum_{k=1}^{K} a_i^{(k)} \mathbf e^{(k)}
\notag\\
&=
\sum_{k=1}^{K} a_i^{(k)}
\underbrace{\mathbf C(\mathbf I-\mathbf P_{\mathbf M})\mathbf e^{(k)}}_{:=\;\varphi^{(k)}\in\mathbb R^3},
\label{eq:supp_delta_pca_expand}
\end{align}

Therefore, the correction term also lies in a low-dimensional subspace of \(\mathbb R^3\), parameterized by the same coefficients \(a_i^{(k)}\). In matrix form, this becomes
\begin{equation}
\delta_i
\approx
\mathbf \Phi\,\mathbf a_i,
\qquad
\mathbf \Phi:=
\bigl[
\varphi^{(1)},\ldots,\varphi^{(K)}
\bigr]\in\mathbb R^{3\times K},
\quad
\mathbf a_i:=
\bigl[a_i^{(1)},\ldots,a_i^{(K)}\bigr]^\top,
\label{eq:supp_delta_lowrank_general}
\end{equation}
where each column is defined as:
\begin{equation}
\varphi^{(k)} := \mathbf C(\mathbf I-\mathbf P_{\mathbf M})\mathbf e^{(k)} \in \mathbb R^{3},
\label{eq:supp_phi_k}
\end{equation}

\paragraph{(5) Rank-1 approximation.}
Empirically, as illustrated in~\cref{supple_fig_2}, the first principal component explains most of the variance of the camera-null spectra. We therefore set \(K=1\) in practice. In this case, the coefficient vector \(\mathbf a_i\) reduces to a scalar \(a_i^{(1)}\), and the calibration matrix \(\mathbf \Phi\) reduces to a single \(3\times 1\) vector \(\varphi^{(1)}\). Hence, \cref{eq:supp_delta_lowrank_general} simplifies to:
\begin{equation}
\delta_i
\approx
\varphi^{(1)} a_i^{(1)},
\label{eq:supp_rank1_delta_compact}
\end{equation}

Using the notation
\begin{equation}
a_i := a_i^{(1)}, \qquad \mathbf e := \mathbf e^{(1)}, \qquad \varphi := \varphi^{(1)} = \mathbf C(\mathbf I-\mathbf P_{\mathbf M})\mathbf e^{(1)},
\label{eq:supp_rank1_notation}
\end{equation}
the rank-1 approximation can be written as:
\begin{equation}
\delta_i
=
\mathbf C(\mathbf I-\mathbf P_{\mathbf M})\mathbf n_i
\approx
\varphi \cdot (a_i\mathbf e),
\label{eq:supp_rank1_delta_final}
\end{equation}
which matches Eq.~(9) in the main paper. Under this interpretation, \(\varphi\in\mathbb R^{3\times 1}\) is a calibratable rank-1 coefficient, while \(a_i\in\mathbb R\) is the scene-dependent estimation coefficient that can be estimated from data by learned predictor.

\paragraph{(6) Interpretation.}
The derivation above shows that the residual correction \(\delta_i\) is governed entirely by the camera-null component of the scene radiance. Once this invisible component is approximated by a low-rank basis, the corresponding correction in camera space is also low-rank. This converts the original high-dimensional and impractical correction problem into the estimation of a small number of coefficients \(a_i\in\mathbb R\) together with a compact calibratable operator \(\varphi\in\mathbb R^{3\times 1}\), which directly motivates the model design in the main paper.

\label{supp_nonlinear_Fc}

\begin{figure}[t]
\centering
  \includegraphics[width=0.9\linewidth]{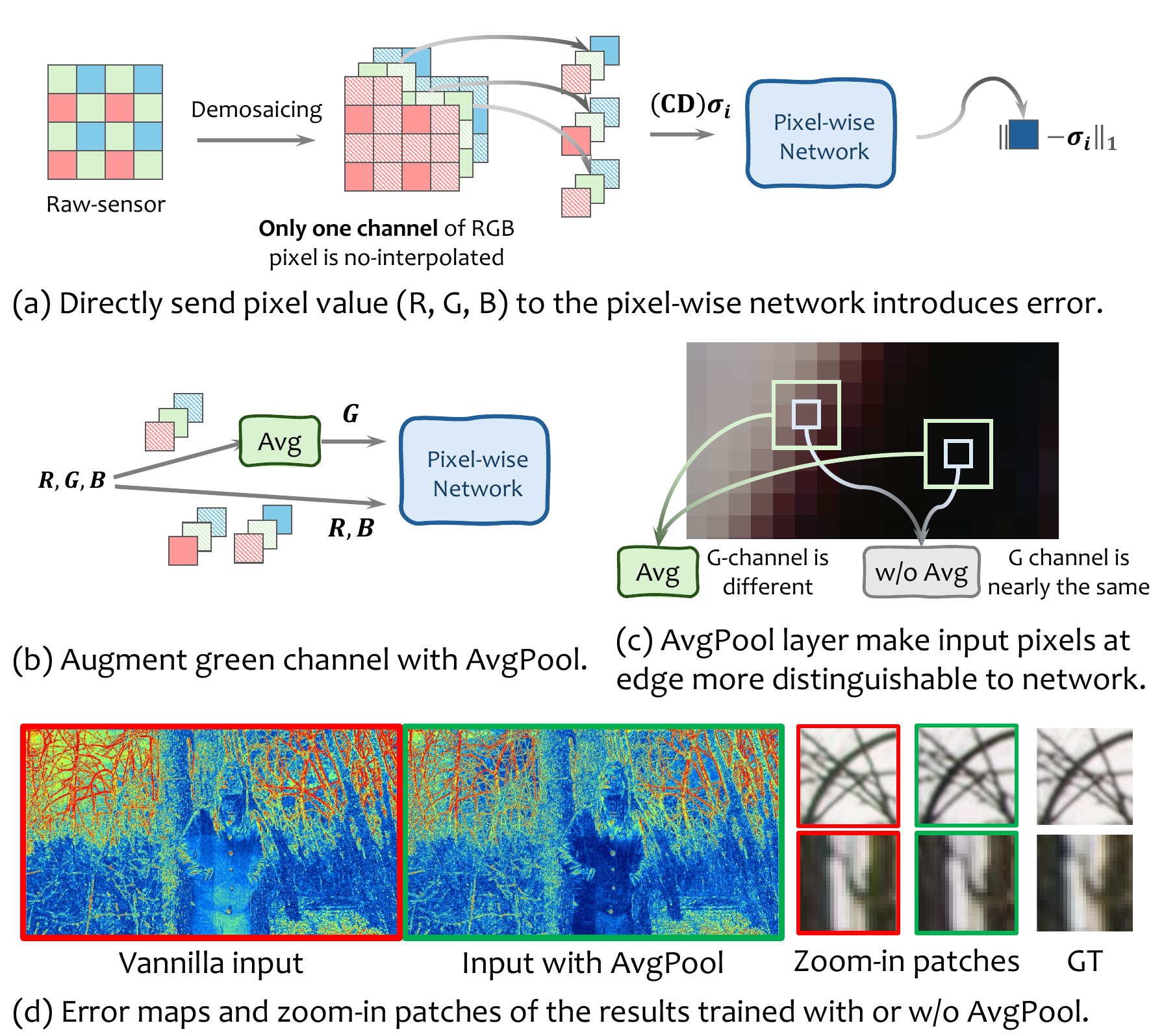}
    \setlength{\abovecaptionskip}{0.3cm}
  \setlength{\belowcaptionskip}{-0.3cm}
    \caption{\textbf{Illustration of the Effectiveness of the AvgPool Layer for Learning \(\mathcal{F}_{\mathbf C}\).} Compared with the vanilla input (a), the input augmented with an AvgPool layer (b) enables the pixel-wise model to distinguish edge pixels introduced by demosaicing (c), thereby producing results with fewer colored edge artifacts (d).}
  \label{supple_fig_5}
\end{figure}

\section{Additional Experiments}
\label{supple-sec-3}
\subsection{Evaluate the Effectiveness of Avgpool Layer for Learning $\mathcal{F}_{\mathbf C}$}
We further analyze the AvgPool layer introduced in Sec.~4.1 for learning the camera--display projector \(\mathcal{F}_{\mathbf C}\). Although \(\widehat{\mathcal{F}}_{\mathbf C}\) is implemented as a pixel-wise mapping, the input camera RGB values are not strictly pixel-local in practice because demosaicing reconstructs each RGB triplet from neighboring sensor samples. Consequently, pixels near edges often contain mixed color responses and are difficult to distinguish from pixels in smooth regions using only per-pixel RGB input. To provide a lightweight spatial cue, we augment the green channel with an AvgPool operation before feeding it to the MLP. This local aggregation helps the model identify pixels affected by demosaicing interpolation, particularly near edges, while preserving the simplicity of the pixel-wise formulation.

As shown in~\cref{supple_fig_5}, compared with the vanilla input, the AvgPool-augmented input makes edge pixels easier for the model to identify and results in fewer colored-edge artifacts in the final output. Empirically, this simple design consistently improves the learning of \(\mathcal{F}_{\mathbf C}\).

\begin{figure}[t]
\centering
  \includegraphics[width=0.9\linewidth]{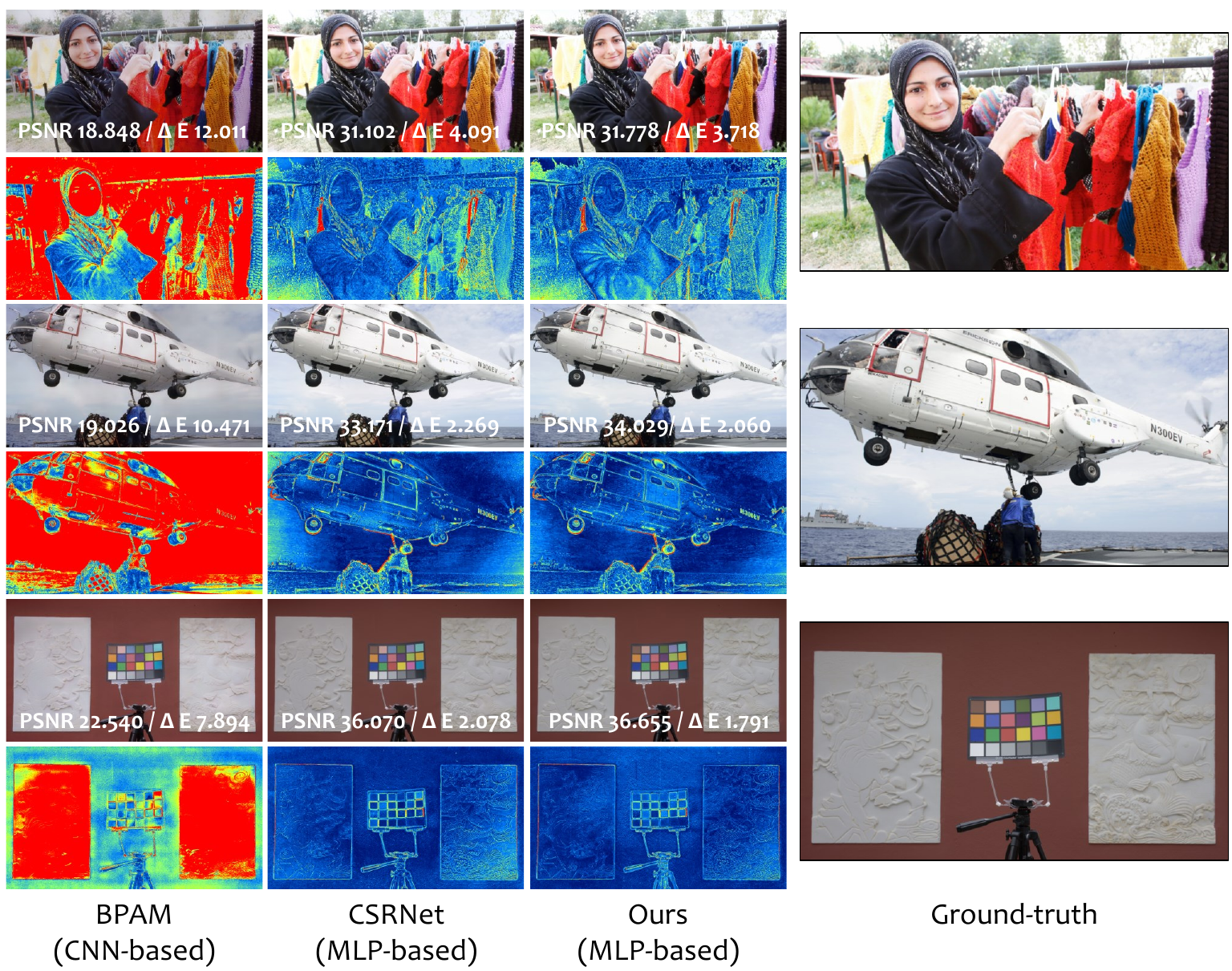}
      \setlength{\abovecaptionskip}{0.3cm}
  \setlength{\belowcaptionskip}{-0.3cm}
    \caption{\textbf{Visualized Fitting Comparisons of \(\mathcal{F}_{\mathbf C}\).} Compared with a CNN-based baseline (BPAM~\cite{lou2025learning}) and an MLP-based baseline (CSRNet~\cite{he2020conditional}), our proposed network yields superior quantitative results and lower errors in the corresponding error maps.}
  \label{supple_fig_6}
\end{figure}

\subsection{Comparison of Different Baselines for Learning \(\mathcal{F}_{\mathbf C}\)}

Tab.~1(a) in the main paper reports the quantitative results for fitting \(\mathcal{F}_{\mathbf C}\) using different baselines, while here we present the corresponding visual comparisons in~\cref{supple_fig_6}. Although BPAM~\cite{lou2025learning}, a CNN-based method, has been shown to significantly outperform CSRNet~\cite{he2020conditional}, an MLP-based method, in color transfer tasks, it does not surpass CSRNet in fitting our camera--display projector \(\mathcal{F}_{\mathbf C}\). This observation suggests that \(\mathcal{F}_{\mathbf C}\) is essentially a pixel-wise color mapping with minimal spatial correlation. Furthermore, our method, which augments an MLP with positional encoding and an AvgPool layer only for the green channel, achieves the best fitting performance with lower error.

\begin{figure}[t]
\centering
  \includegraphics[width=0.9\linewidth]{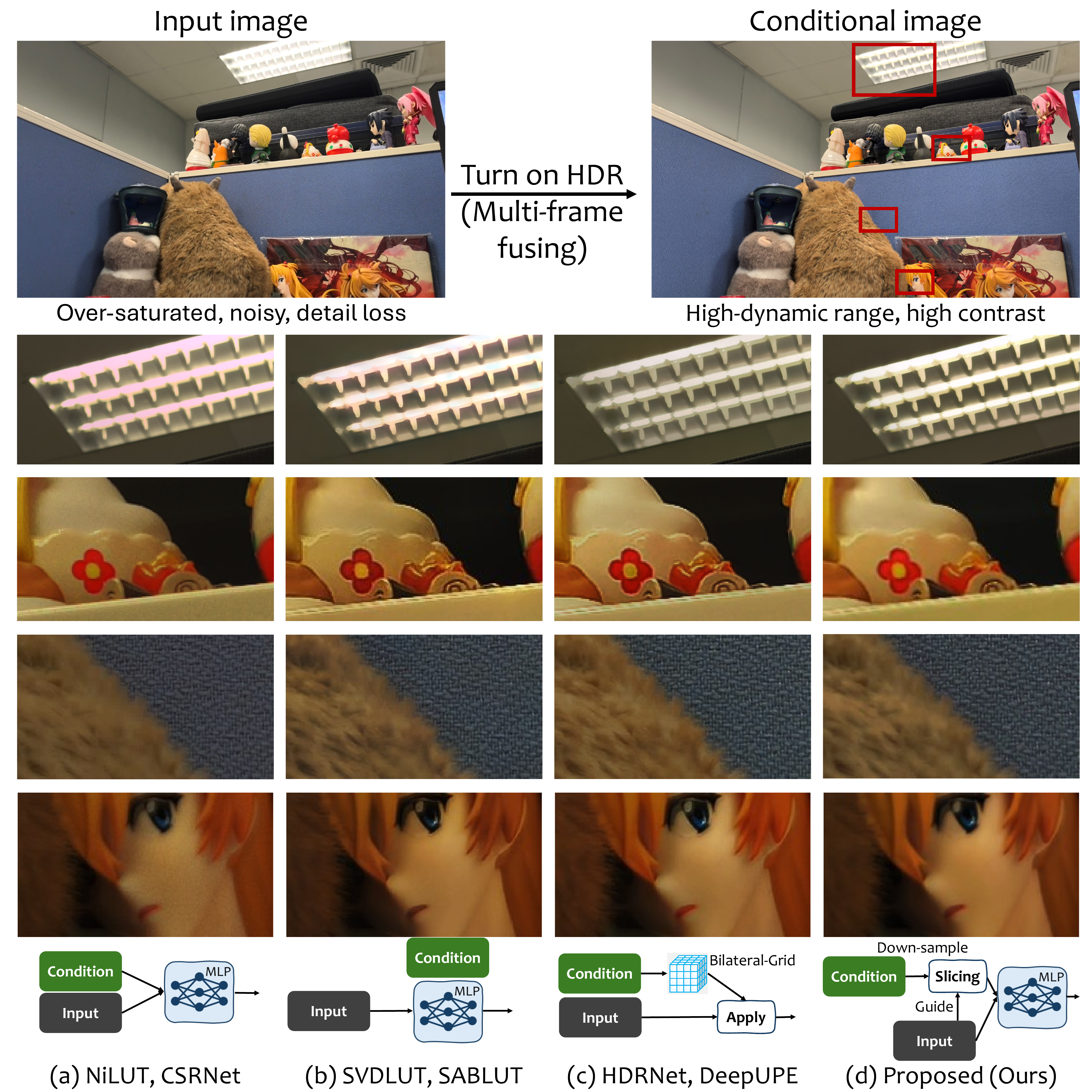}
        \setlength{\abovecaptionskip}{0.3cm}
  \setlength{\belowcaptionskip}{-0.3cm}
    \caption{\textbf{Comparison of ISP-Augmented Color Pass-through.} In addition to the original input image (top left), we use the burst-fusion image as a condition (top right) to augment Color Pass-Through. We compare designs inspired by (a) NiLUT~\cite{conde2024nilut} and CSRNet~\cite{he2020conditional}, (b) SVDLUT~\cite{kim2025lightweight} and SABLUT~\cite{kim2024image}, (c) HDRNet~\cite{gharbi2017deep} and DeepUPE~\cite{wang2019underexposed}, and (d) our design. Our method produces results with natural highlights (first row), vivid colors (second row), richer details (third row), and lower noise (fourth row).}
  \label{supple_fig_7}
\end{figure}

\section{Further Application with Color Pass-through}
\label{supple-sec-4}
\subsection{Augment with In-phone Camera Processing}
The proposed Color Pass-Through model currently supports only single-frame input with relatively simple post-processing. Here, we discuss a possible extension that integrates it with commercial image signal processing (ISP) pipeline.

Modern smartphones typically include built-in imaging functions such as HDR~\cite{Debevec1997HDR}, burst fusion~\cite{Hasinoff2016BurstMobileHDR}, and tone mapping~\cite{Reinhard2002ToneMapping}, which improve dynamic range, detail, and contrast. Motivated by this, we use the HDR-fused image as a \textbf{condition} while keeping the original raw image as the \textbf{input}, as illustrated in~\cref{supple_fig_7}. Among several tested designs, the variant in~\cref{supple_fig_7}(d) yields the best results.

\subsection{Extension to High Dynamic Range}
\label{sec-Extension_to_High_Dynamic_Range}
\begin{figure}[t]
\centering
  \includegraphics[width=0.95\linewidth]{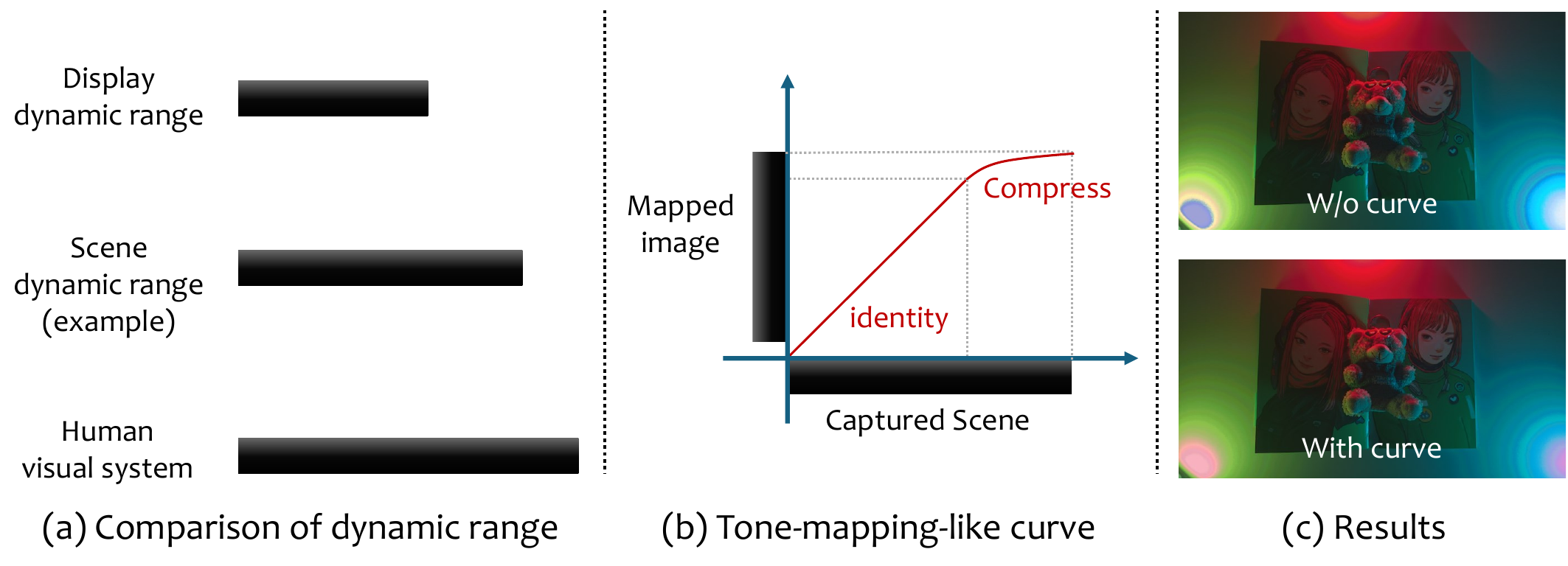}
        \setlength{\abovecaptionskip}{0.3cm}
  \setlength{\belowcaptionskip}{-0.3cm}
    \caption{\textbf{Comparison of Curve-Augmented Color Pass-Through.} 
    (a) Because displays have a more limited dynamic range than real scenes, the learned mapping may fail in out-of-range regions. 
    (b) To mitigate this issue, we apply a tone-mapping-like curve to the input image, using identity mapping in low-intensity regions and compression in high-intensity regions. 
    (c) Visual comparison with and without the proposed curve. The augmented version produces more consistent color variations in strongly illuminated regions (e.g., the bottom corners).}
  \label{supple_fig_8}
\end{figure}

The proposed Color Pass-Through model is designed to \textbf{strictly} align the display brightness with that of the real-world scene, since our goal is for the display to try it best to faithfully \textbf{reproduce} the scene color.

However, due to hardware limitations, current displays still fall far short of the upper range of human visual perception, especially in scenes containing strong highlights (see~\cref{supple_fig_8}(a)). Here, we explore a simple extension that augments the original input with a tone-mapping-like curve in order to better preserve high-intensity highlight regions (see~\cref{supple_fig_8}(b)). As shown in~\cref{supple_fig_8}(c), this strategy can alleviate overexposure to some extent, although a more effective solution remains an important direction for future investigation.

% \subsection{Generalize to Different Camera}

\section{Limitation}
\label{supple-sec-5}
\paragraph{Generalization across different cameras.}
Our Color Pass-Through framework is designed for a specific camera--display pair, such as Vision Pro reproducing a real scene captured by a particular camera through its display. As a result, the learned model cannot be directly applied to images captured by a different camera. A possible direction to address this limitation is raw-to-raw translation~\cite{Gronquist2025BeyondCalibration,Perevozchikov2024Rawformer,Xie2024GeneralizingISP}, which could potentially be combined with our full model to enable cross-device transfer in future work.

\paragraph{Limited dynamic range.}
As discussed in~\cref{sec-Extension_to_High_Dynamic_Range}, current displays have a limited dynamic range compared with real-world scenes. Under extreme illumination conditions, such as direct sunlight, faithful color reproduction may therefore require HDR capture, specific tone-mapping techniques and rendering to avoid clipping or compression in bright regions.

\end{document}